\documentclass[journal]{IEEEtai}

\usepackage[colorlinks,urlcolor=blue,linkcolor=blue,citecolor=blue]{hyperref}

\usepackage{color,array}
\usepackage{balance} 
\usepackage{graphicx}

\usepackage{ulem}
\usepackage{afterpage}  
\usepackage{multirow}
\usepackage{booktabs} 
\usepackage{amssymb}
\usepackage{relsize}
\usepackage{amsmath} 
\usepackage{amsfonts}
\usepackage{hyperref}
\usepackage{supertabular}
\usepackage{xltabular}
\usepackage[switch]{lineno} 
\usepackage{longtable}
\usepackage[T1]{fontenc}
\newtheorem{definition}{Definition}[section]
\usepackage{multirow}
\usepackage{balance}
\usepackage{mathrsfs}
\usepackage{svg}
\usepackage{multirow}
\usepackage{caption}
\usepackage{pifont}
\usepackage{float}
\usepackage{soul}
\usepackage{url}
\usepackage[utf8]{inputenc}
\usepackage{graphicx}
\usepackage{amsmath}
\usepackage{booktabs}
\usepackage{subfigure}
\usepackage{booktabs,tabularx}
\usepackage{amsfonts}
\usepackage{extarrows}
\usepackage{color}
\usepackage{enumerate} 
\usepackage{algorithmic}  
\usepackage{verbatim}
\usepackage{rotating}
\usepackage{hyperref}
\usepackage{wrapfig}
\usepackage{cases}
\usepackage{xcolor}
\usepackage{ltablex}
\keepXColumns

\DeclareMathOperator*{\argmin}{arg\,min}

\newcolumntype{Y}{>{\raggedright\arraybackslash}p{2.8cm}}
\newcolumntype{Z}{>{\raggedright\arraybackslash}p{2.8cm}}

\IEEEoverridecommandlockouts

\begin{document}

\title{Explainable AI for Mental Disorder Detection on Social Media: A Survey and Outlook} 

\author{Yusif Ibrahimov, Tarique Anwar, Tommy Yuan
\thanks{Yusif Ibrahimov is with 
the Department of Computer Science, University of York, Heslington, United Kingdom (e-mail:yusif.ibrahimov@york.ac.uk) and French-Azerbaijani University under Azerbaijan State Oil and Industry University (e-mail:yusif.ibrahimov@ufaz.az)}
\thanks{Tarique Anwar is with 
the School of Computing Technologies, RMIT University, Melbourne, Australia (e-mail: tarique.anwar@rmit.edu.au).}
\thanks{Tommy Yuan is with 
the Department of Computer Science, University of York, Heslington, United Kingdom (e-mail: tommy.yuan@york.ac.uk).}
}


\IEEEpubid{
    \makebox[\columnwidth]{
        \begin{tabular}{p{\columnwidth}}
            \textcopyright 2026 IEEE. Personal use of this material is permitted. Permission from IEEE must be obtained for all other uses, in any current or future media, including reprinting/republishing this material for advertising or promotional purposes, creating new collective works, for resale or redistribution to servers or lists, or reuse of any copyrighted component of this work in other works.
        \end{tabular}
    }
    \hspace{\columnsep}\makebox[\columnwidth]{}
}

\maketitle

\begin{sloppypar}

\IEEEpubidadjcol 

\begin{abstract}

Mental health is a complex global challenge. The widespread use of online social media (OSM) provides a rich source of data for supporting mental health analytics. This paper presents a {comprehensive} review of AI- and data-driven approaches for mental disorder detection using OSM, with a particular focus on explainable artificial intelligence (XAI). We summarise traditional diagnostic practices and provide an integrated overview of modern deep learning models, emphasising how their growing complexity increases the need for transparent and trustworthy explanations. We review a wide range of explainability techniques applied in this domain, including interpretable models, post-hoc attribution methods, attention- and graph-based explanations, and emerging LLM-driven approaches, and outline commonly used datasets and evaluation practices. Finally, we identify key challenges such as data limitations, bias, and the need for clinically meaningful explanations, and propose directions for future research. This survey offers a comprehensive foundation for developing explainable, reliable, and ethically informed mental disorder detection systems for social media.

\end{abstract}

\begin{IEEEImpStatement}
Mental disorders impose a significant global burden. As online social media platforms increasingly serve as spaces where individuals express their mental states, AI-driven analysis of these data offers opportunities for scalable, real-time mental health support. However, the opacity of modern AI models introduces risks of bias, misuse, and reduced clinical trust. This survey provides a comprehensive, explainability-focused synthesis of mental disorder detection from social media, integrating insights from feature engineering, predictive modelling, domain knowledge, and explainability techniques. It also identifies key research challenges and outlines directions for developing transparent, trustworthy, and clinically grounded AI systems for mental healthcare.\end{IEEEImpStatement}


\section{Introduction} \label{sec:introduction}

\IEEEPARstart{M}{ental} health (MH) influences psychological well-being, affective states, and behavioural patterns. Mental disorders represent a major global public health challenge, affecting about 12.5\% of individuals during their lifetime and nearly 1 billion people worldwide \cite{rehm2019global}. Conditions such as depression, anxiety, bipolar disorder, and schizophrenia impose substantial social, healthcare, and economic burdens \cite{ZHANG2023231, who_mental_disorders}. They are also associated with serious health outcomes, including self-harm, cardiovascular disease, cancer, and suicide, which accounts for around 800,000 deaths annually worldwide \cite{rehm2019global, who_mental_disorders,10.1145/3572406}. Despite their prevalence, many individuals delay or avoid seeking professional support due to stigma and structural barriers \cite{Thompson2004, henderson2020mental}. Treatment gaps remain severe, particularly in low-income countries where only 1 in 27 individuals receives adequate care \cite{thornicroft2017undertreatment}. Consequently, untreated conditions lead to significant functional impairment and economic costs, estimated at USD 2.5 trillion globally in 2010 \cite{trautmann2016economic}. These challenges are further intensified during crises such as pandemics and armed conflicts, which disrupt access to mental health services \cite{10.1145/3543507.3583863,9447025}.

Addressing this global challenge requires innovative strategies for early identification, continuous monitoring, and personalised intervention. In recent years, the intersection of data science, machine learning (ML), artificial intelligence (AI), and mental healthcare has offered promising avenues for such solutions \cite{Zhang2022-gm}. AI-driven approaches support early diagnostics \cite{10.1145/3485447.3512128}, therapy \cite{DALFONSO2020112}, and monitoring \cite{10.1145/3471902}, potentially alleviating individual suffering and reducing healthcare, societal, and economic burdens. Online social media (OSM) platforms constitute a rich source of personal data, with approximately 60\% of the global population engaging actively \cite{datareportal_social_media_users, anwar2023tracking}. Particularly among youth, sensitive topics, emotions, and daily experiences are frequently expressed online rather than in-person \cite{10.1145/3539618.3591905}. These digital traces generate a vast repository of behavioural and emotional data with strong potential for mental health analytics \cite{9447025,bucur2021psychologically,coppersmith-etal-2016-exploratory}. Recent studies highlight OSM as a promising avenue for continuous mental healthcare \cite{Sanchez2020-wf}.

Deep learning (DL) models \cite{10.1145/3404835.3462938,10.1145/3543507.3583863} have demonstrated strong predictive performance for detecting mental disorders from OSM data. However, their increasing complexity poses challenges for adoption in clinical settings due to their ``black-box'' nature \cite{Zogan2022-wd}. Relying on opaque models in healthcare raises ethical and safety concerns, as even high-performing models cannot guarantee perfect accuracy \cite{9769937}. Explainable AI (XAI) addresses these challenges by elucidating the decision-making processes of AI models, promoting transparency, trust, and accountability in sensitive applications. Integrating XAI into deep learning models ensures that predictions are interpretable, actionable, and aligned with clinical needs. Given the growing adoption of AI in mental healthcare, investigating deep learning models with an emphasis on explainability is critical. By combining predictive power with transparency, XAI facilitates safer, ethically responsible, and clinically meaningful mental disorder detection systems. 

\subsection{Existing surveys vs our contributions}

Several surveys have examined AI-based mental disorder detection (MDD) using OSM data~\cite{thieme_rev,10108975,9568643,Zhang2022-gm,schonning2020social,chancellor2020methods}. However, most are limited in scope, technical depth, or focus on explainability. For example, Hasib et al.~\cite{10108975} and Yu et al.~\cite{9568643} focus on specific disorders (e.g., depression or suicidality), limiting generalisability. Zhang et al.~\cite{Zhang2022-gm} include OSM as a data source but discuss platform-specific characteristics only briefly and provide limited analysis of explainability. Chancellor et al.~\cite{chancellor2020methods} review predictive methods but provide little discussion of model architectures, domain-specific evaluation strategies, or explainability. Similarly, Thieme et al.~\cite{thieme_rev} offer a broad overview of machine learning in mental health but treat OSM only as a subdomain, with limited attention to its unique characteristics and challenges.

In contrast, our survey provides a comprehensive and technically detailed review of explainable MDD using OSM data. Our main contributions are:

\begin{itemize}

\item[-] \textit{Feature extraction with explainability in mind}: We present a taxonomy of features relevant to MDD, including linguistic cues, posting behaviours, temporal patterns, and social network structures, along with domain-informed features derived from psychological theory.

\item[-] \textit{Comprehensive review of predictive models}: We examine classical machine learning and deep learning approaches, comparing their strengths, limitations, and suitability for mental health, with attention to explainability.

\item[-] \textit{Clinical grounding through domain knowledge}: We discuss how knowledge graphs, psycholinguistic evidence, clinical questionnaire modelling, and domain-specific embeddings can anchor model explanations in validated mental-health constructs.

\item[-] \textit{In-depth exploration of explainability methods}: We analyse explainability techniques including LIME, SHAP, attention-based mechanisms, and emerging LLM-based explanations, and highlight underused approaches such as GNN explainability, ICE, and PDP.

\item[-] \textit{Datasets and evaluation}: We summarise key datasets used for MDD on OSM and review evaluation approaches beyond standard predictive metrics, including explainability-oriented and domain-specific measures.

\item[-] \textit{Challenges and future directions}: We synthesise major technical, ethical, and methodological challenges and outline research directions for developing trustworthy and explainable MDD systems.
\end{itemize}

Our survey is guided by four research questions:

\begin{enumerate}
\item[\textbf{Q1}] What are the limitations and challenges of traditional methods for diagnosing mental disorders?
\item[\textbf{Q2}] What are the advantages and disadvantages of AI-driven approaches for detecting mental disorders via OSM?
\item[\textbf{Q3}] What is the current state of explainability in AI-driven MDD models?
\item[\textbf{Q4}] What are the challenges and potential future research directions in explainable AI for MDD?
\end{enumerate}

\begin{figure}[t]
    \centering
    \includegraphics[width=0.5\textwidth]{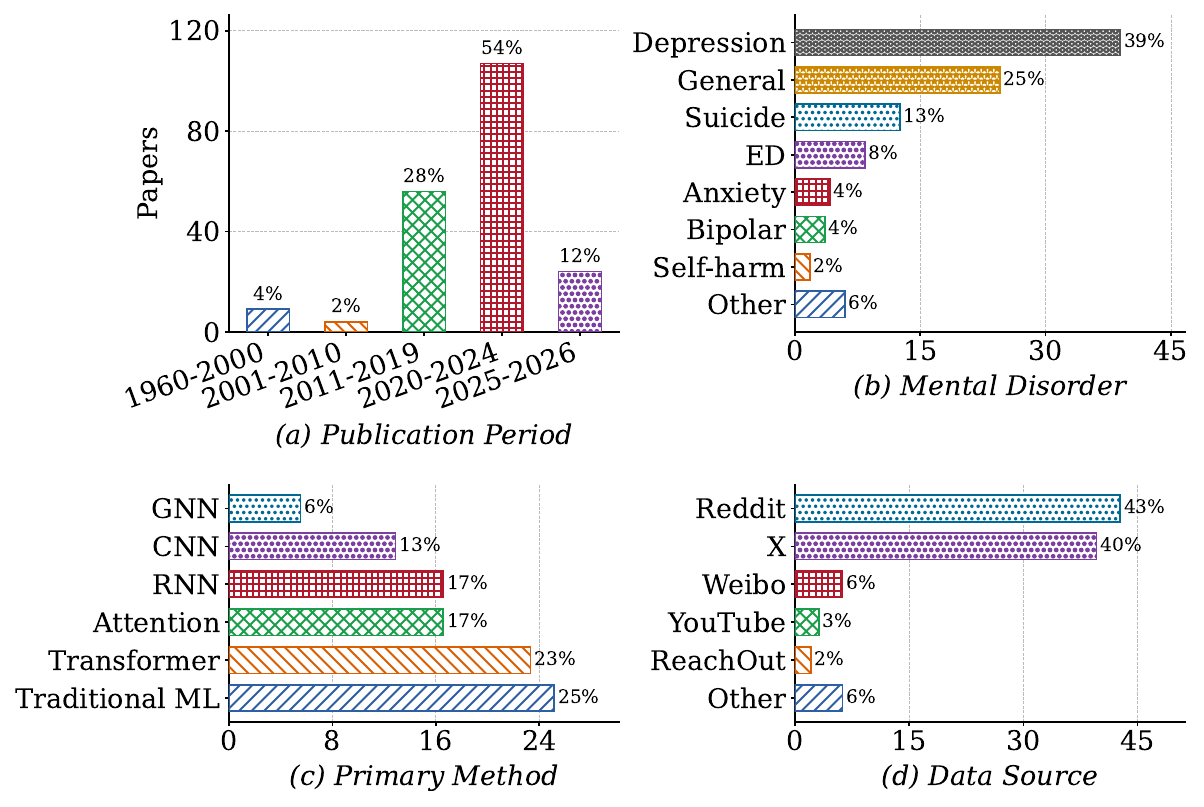}
    \caption{{Bibliometric overview of referenced paper.}}
    \label{fig:bibliometric}
\end{figure}

{
\subsection{Survey methodology and overview}

For transparency, this section outlines the literature search and selection procedure for this comprehensive narrative survey\footnote{{A previous version of this manuscript was made publicly available on arXiv (\url{https://arxiv.org/abs/2406.05984}). The current version incorporates substantial revisions, including expanded coverage of contemporary XAI methods and ethical considerations.}}. We consulted major indexes and databases including DBLP, Scopus, Web of Science, and PubMed, using keywords related to mental health, OSM, ML, and XAI. The search began with state-of-the-art publications in leading venues, prioritising studies at the intersection of XAI, social media analytics, and mental disorder detection. We expanded the coverage via backward and forward snowballing of references and citations, capturing influential and emerging works. Our selection criteria focused on mental health analysis using OSM data, relevance to explainable or interpretable modelling, clinical or domain relevance, and methodological quality. This approach supports a structured yet flexible synthesis of methods and insights, which is appropriate for interdisciplinary AI surveys where interpretative comparison and taxonomy development are essential.

Figure~\ref{fig:bibliometric} presents a bibliometric overview of the reviewed literature. More than 75\% of the surveyed studies were published between 2020 and 2026, reflecting the rapid growth of this research area. Depression, general mental health conditions, and anxiety are the most frequently studied disorders, partly due to the clearer linguistic signals observable in social media discourse. Over 80\%  of the studies utilise Reddit and $\mathbb{X}$, largely due to the relative accessibilityof their data and APIs. Traditional ML models and transformer-based architectures are the most commonly adopted computational approaches, reflecting the balance between interpretability and predictive performance in the literature.
}

\subsection{Organisation}
The remainder of the paper is organised as follows. Section~\ref{sec:traditionalmethods} reviews traditional diagnostic approaches for mental disorders. Section~\ref{sec:datadrivenmethods} surveys XAI techniques for MDD, including feature extraction, ML/DL models, the role of domain knowledge, and explainability methods. Section~\ref{sec:experimentaldesign} summarises commonly used datasets and evaluation practices. Section~\ref{sec:researchhorizons} discusses key challenges and future research directions, followed by concluding remarks in Section~\ref{sec:conclusion}.

\section{Traditional Diagnostic Methods}
\label{sec:traditionalmethods}

Mental health conditions are typically diagnosed by healthcare professionals (e.g. therapists and psychologists) through face-to-face interviews with patients. The patients may be required to respond to pre-defined standard questionnaires to enhance the quality of diagnosis, with each type of mental disorder often having its own set of self-questionnaires \cite{Beck1961-qd, doi:10.1177/014662167700100306, Cooper1987-ov, garner_garfinkel_1979}. This comprehensive diagnostic process is guided by individuals' experiences, with international standards such as the Diagnostic and Statistical Manual of Mental Disorders (DSM-5) employed for diagnosis \cite{9447025,DSM5}. DSM-5 \cite{DSM5} is a manual listing all mental disorders, along with their standard diagnostic methods. Main symptoms of various mental disorders are identified through these methods. For example, depression is characterised by a loss of interest in previously enjoyed activities, feelings of hopelessness, reduced motivation, concentration, energy and libido, and changes in cognition \cite{DSM5}. Anxiety manifests through symptoms like tachycardia, dizziness, shortness of breath, trembling, worry, anger, and fear \cite{DSM5}. Eating disorders are identified through signs such as problematic eating habits, abnormal changes in weight, changes in mood and unhappiness with body shape \cite{DSM5}.
Several standardised questionnaires play a pivotal role in assessing mental conditions. The depression inventory by Beck et al. \cite{Beck1961-qd} and the CES-D Scale \cite{doi:10.1177/014662167700100306}, each comprising 21 and 20 questions, respectively, assess the mental status for depression diagnosis. Gold standards for estimating depression severity include the Beck Depression Index-II (BDI-II) \cite{beck_ii}, Hamilton Depression Rating Scale (HDRS) \cite{10.5555/3294771.3294869}, and Patient Health Questionnaire-9 (PHQ-9) \cite{Kroenke2001-ul}, focusing on key depression symptoms such as sadness, pessimism, loss of interest, and fatigue. Recognised questionnaires for anxiety detection include the Generalised Anxiety Disorder 7 (GAD-7) \cite{10.1001/archinte.166.10.1092} and Beck Anxiety Disorder \cite{Beck_undated-xw}. For the assessment and diagnosis of eating disorders (EDs), the Eating Disorder Examination Questionnaire (EDE-Q) \cite{Cooper1987-ov} and the Eating Attitudes Test (EAT-26) \cite{garner_garfinkel_1979} are employed. 

Despite the well-established area of mental healthcare relying on traditional diagnostic methods, there are inherent limitations in the current approaches \cite{10.1145/3589784, KADKHODA2022101042} -- $(i)$ The \textit{frequency of monitoring sessions} is typically limited to once per week. It hinders the ability to track the course of mental conditions in real-time. $(ii)$ The \textit{subjective approach} of clinicians may lead to instances of subjective judgment, introducing potential inaccuracies in diagnosis . $(iii)$ \textit{Accessibility to therapy sessions} may be negatively affected by factors such as financial constrains and societal circumstances, particularly during crisis like pandemics and wars. $(iv)$ \textit{The phenomenon of patient prejudice} is seen quite often, as patients' responses to self-questionnaires may exhibit bias due to individual interpretations, contextual factors, and potential misunderstandings of the questions, including intentional concealment of information. $(v)$ Additionally, \textit{humanity barriers}, including cultural diversity, can pose challenges for patients in effectively communicating their needs due to language and traditional limitations.

For a timely detection of mental disorders while addressing the existing limitations, it is crucial to explore widely accessible, impartial, consistently monitored, and immediately available alternatives. As mentioned earlier, the ubiquitous OSM platforms are used by a vast majority of individuals, and therefore serve as a repository of substantial volumes of individual data. To address the limitations observed in conventional diagnostic approaches, this study aims to examine preliminary diagnostic methods for MDD using AI-driven techniques. 

\begin{table*}[!h]
\caption{Some common mental disorders, their definitions, and indicative examples of OSM posts}
\label{tab:data_example}
\centering
\begin{tabularx}{\textwidth}{|p{2.5cm}|X|X|}
\hline
\textbf{Mental disorder} & \textbf{Definition} & \textbf{Example of OSM post} \\
\hline
Depression & Negative changes marked by affect, cognition, mood, and neurovegetative functions lasting at least 2 weeks \cite{DSM5,9447025}. & \textit{Can Someone Cheer Me Up? - I'm diagnosed with depression. I will sometimes out of nowhere...} \\
    \hline
    Anxiety & Conditions characterised by excessive fear and anxiety, along with associated mental and behavioral disorders \cite{DSM5, Richter2020-jn}. & \textit{Anxiety is seriously affecting my life and it is ruining my life and I don't know how I'm going to provide for myself in the future} \\
    \hline
    Eating Disorder & A persistent disturbance in eating behavior that leads to altered food consumption and significantly impairs mental and behavioral functioning \cite{DSM5, 10.1145/3543507.3583863}. & \textit{Getting treatment for an eating disorder is awful and I wish I never started it. (Vent / cry session haha)} \\
    \hline
    Bipolar Disorder & Abnormally and persistently elevated, expansive, or irritable mood, along with increased activity or energy \cite{DSM5, KADKHODA2022101042}. & \textit{Finally after years long struggle I considered seeking a psychiatrist. And just diagnosed bipolar disorder. She prescribed me with two medicines} \\
    \hline
    Schizophrenia & Abnormalities in delusions, hallucinations, disorganised thinking, and grossly disorganised or abnormal motor behavior \cite{DSM5, mitchell-etal-2015-quantifying}. & \textit{Hi. I’m **** and I’m a ********* with schizophrenia and I take a medication. I also have evil hallucinations and it scares me. Is it my schizophrenia or is it really the Devil?} \\
    \hline
    Suicide Ideation & Thoughts or contemplation of taking one's own life or self-inflicted harm \cite{DSM5,9308975}. & \textit{I'm having a really hard time with a lot of things in my life right now. My father started to verbally abuse me and my sisters since I was 6 or 7. I want to commit suicide.} \\
    \hline
    Obsessive-Compulsive Disorder (OCD) & Characterised by the presence of recurrent and persistent thoughts, urges, or images, along with repetitive behaviors or mental acts \cite{DSM5, doi:10.1177/0276236615598957}. & \textit{I've been going to a doctor to see if I had autism and while there they also diagnosed me with OCD.} \\
\hline
\end{tabularx}
\end{table*}

\section{Explainable AI for Mental Disorder Detection on Social Media} 
\label{sec:datadrivenmethods}

Text-based OSM platforms such as $\mathbb{X}$ and Reddit provide rich signals for MDD, including users’ posts, behavioural patterns, and social interactions. These platforms capture everyday expressions of mood, cognition, and distress, making them valuable resources for mental health analysis. Table \ref{tab:data_example} presents common mental disorders. While AI and deep learning models trained on such data achieve strong predictive performance, their internal decision processes often remain opaque. In clinical contexts, this opacity is problematic, as decisions influencing risk assessment or intervention require transparency and alignment with psychological constructs. Explainable AI is therefore essential to interpret how models use social signals and to ensure predictions can be safely and ethically integrated into mental healthcare.

At a high level, XAI methods can be categorised into \textit{interpretable models} and \textit{post-hoc explainability methods} \cite{Jia2022-hp}. Interpretable models are inherently transparent, allowing users to understand their internal decision-making directly (e.g., decision trees, linear or logistic regression). In contrast, complex DL models, typically require post-hoc methods to generate explanations after training. Post-hoc explainability methods are commonly categorised along two dimensions: \textit{scope} and \textit{dependence on the model} \cite{Arya2019-yg, Jia2022-hp, Yuan2022-lk, Danilevsky2020-cv}. \textit{Scope} distinguishes between \textit{local explainability}, which explains individual predictions, and \textit{global explainability}, which provides insights into the model’s overall decision-making process. \textit{Dependence on the model} differentiates \textit{model-agnostic methods}, applicable to any trained model, from \textit{model-specific methods}, tailored to particular architectures.

\begin{figure}[!ht]
  \centering
  \includegraphics[width=0.48\textwidth]{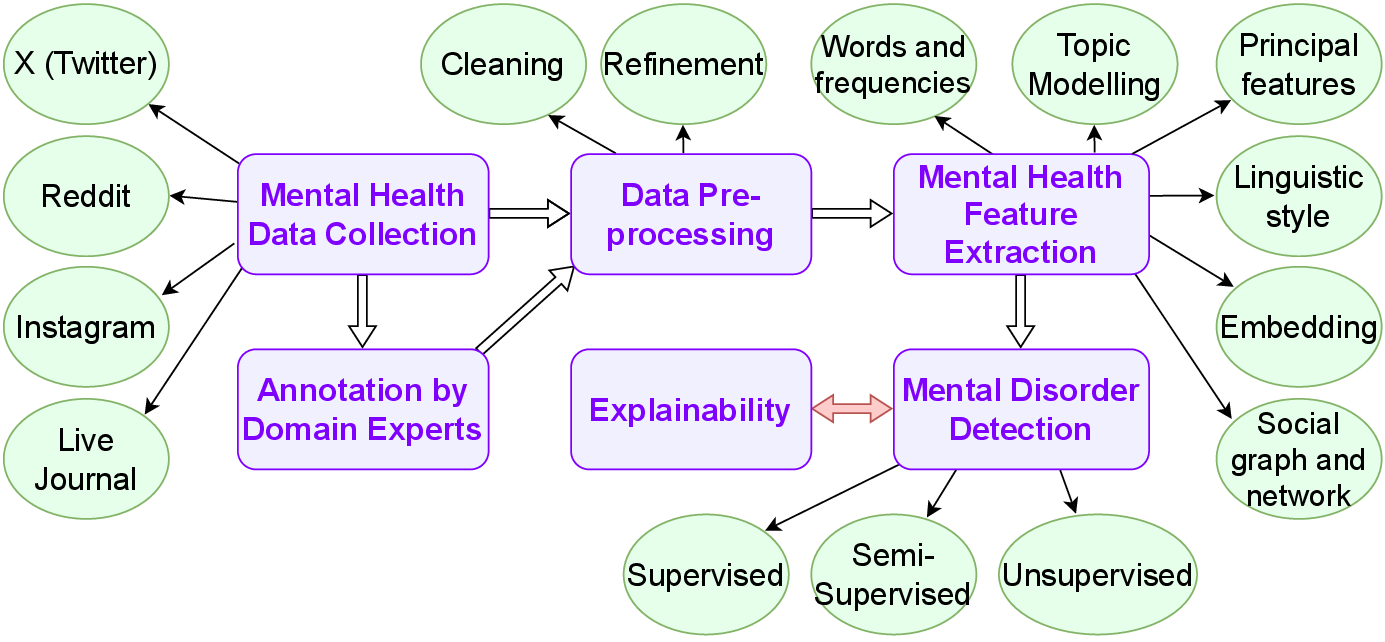}
  \caption{Workflow of explainable AI methods for MDD}
  \label{fig:gen_workflow}
\end{figure}

\begin{figure*}[!ht]
  \centering
  \includegraphics[width=1\textwidth]{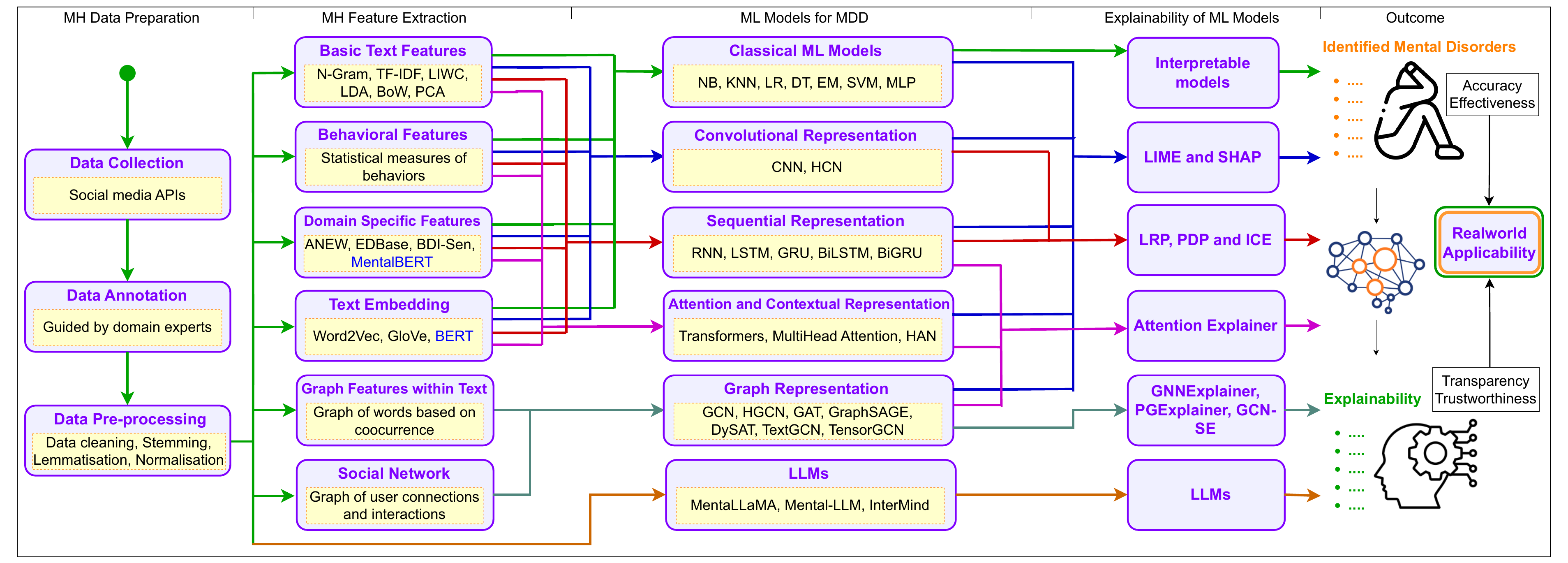}
  \caption{Explainable AI pipeline for MDD via social media}
  \label{fig:workflow}
\end{figure*}

Figure \ref{fig:gen_workflow} illustrates the general workflow for MDD using OSM data. Each stage provides opportunities to introduce transparency. The process begins with collecting mental health–related data from OSM platforms and pre-processing for preparation. Feature extraction then transforms the cleaned data into quantified and structured representations (Section \ref{sec:featureextraction}). The modelling stage constructs systems capable of identifying patterns associated with mental disorders. Approaches include supervised, unsupervised, and semi-supervised learning, although supervised models remain the most widely adopted (Section \ref{sec:mlmdd}). These models form the core of the predictive pipeline but also introduce opacity, highlighting the need for explainability. The final stage focuses on generating explanations for model behaviour (Section \ref{sec:explainability}). Figure \ref{fig:workflow} expands this workflow into a detailed XAI pipeline, illustrating where different explainability techniques integrate with the modelling process. 

{Although MDD studies primarily rely on textual signals from social media, multimodal data are increasingly being explored. For example, Al Jazaery and Guo \cite{al2018video} used CNNs and RNNs to estimate BDI-II values from video data. Yoon et al. \cite{yoon2022d} created the D-Vlog dataset of 961 YouTube vlogs, fusing visual and acoustic features via attention mechanisms. Li et al. \cite{li2025depressinstruct} developed a multimodal instruction dataset for fine-tuning large language models (LLMs) that integrates signals, transcriptions, and emotions. Zhang et al. \cite{zhang2026sounding} combined speech and text data for personalized depression detection using eGeMAPS acoustic features and BERT text embeddings. These approaches illustrate that multimodal data can reveal richer mental disorder indicators, motivating broader multimodal research in MDD.
}

\subsection{Mental health feature extraction} \label{sec:featureextraction}

Online social media data on mental health include unstructured text, metadata, behavioural patterns, and social networks. Structuring these heterogeneous signals into meaningful features is essential for ML/DL models to detect psychological patterns and support interpretable modelling. The nature of the extracted features greatly influences what can later be explained.



\subsubsection{Basic text representations}
Methods such as bag of words (BoW), TF-IDF, and n-grams \cite{10.1145/3136755.3136766,7752261,7752434,Yan2019-ih,Shickel2016-su,10.1145/3442536.3442553,10.1145/3543507.3583350,9918782,9416889,hameed2025explainable} form the foundation of text representation for mental health analysis. 
Topic models such as latent dirichlet allocation (LDA) \cite{Ma2023-ts,Hagg2022-tb} uncover latent themes in user language, while dimensionality reduction methods like PCA \cite{Lee2020-ps,Tlachac2020-qw} help reduce sparsity. Psycholinguistic tools such as LIWC \cite{Coppersmith2014-kw,Tlachac2020-qw,10.1145/3442536.3442553} capture emotional, cognitive, and social indicators highly relevant to MDD.
These feature types are relatively transparent, which makes them amenable to interpretable and post-hoc XAI methods.


\subsubsection{Deep embeddings}
Deep word embedding methods capture semantic and contextual information beyond traditional feature engineering. 
Word2Vec \cite{Mikolov2013-kq} and GloVe \cite{Pennington2014-fc} are widely used in MH studies, where Word2Vec learns distributed word representations and GloVe captures broader semantics \cite{Straw2020-lp,Liu2022-qx,Ahmed2022-oy,Dheeraj2021-at,10.1145/3442536.3442553,hameed2025explainable}.
Transformer-based models such as BERT \cite{Devlin2018-zf} produce contextualised embeddings that reflect the linguistic and emotional nuances of user posts \cite{10.1145/3404835.3462938,bayram-benhiba-2022-emotionally}.
While powerful, these representations increase model opacity, making XAI techniques essential for interpreting the resulting deep models.

\subsubsection{Behavioural features}

Users’ behaviour on OSM provides important signals beyond text content. Features such as posting frequency, diurnal activity rhythms, interaction rates, and network engagement \cite{9447025,10.1145/3543507.3583863,10.1145/3404835.3462938} often correlate with mental health states. Behavioural indicators complement textual cues and contribute to more holistic modelling.
Because these features correspond to observable user patterns, they are often easier to interpret and integrate into explainability frameworks.


\subsubsection{Domain-specific features}
Domain-specific lexical and model-based features incorporate clinically grounded semantics. Lexicons such as ANEW \cite{Bradley1999AffectiveNF} and EDBase \cite{9906413} encode affective or disorder-related terminology, enriching linguistic analysis with psychological meaning. Domain-adapted language models like MentalBERT \cite{Ji2021-hu} further tailor representations to mental health contexts.
These features make explanations more clinically meaningful, as they align directly with constructs recognised by mental health professionals.

\subsubsection{Social network representation} 
OSM platforms connect individuals through complex social networks \cite{datareportal_social_media_users}. Research suggests a correlation between an individual's mental health and that of their friends, highlighting the potential spread of mental health issues within peer networks \cite{Rosenquist2011-jx, alho2024transmission, kiuru2012depression}. 
Therefore, the infusion of social network features into individual-level features can benefit MDD \cite{10027637,kuo2023dynamic}. The social network features also support explainability by linking predictions to peer influence patterns or structural network properties.

\subsection{Machine learning on social media data for MDD}
\label{sec:mlmdd}
The task of MDD from OSM data poses challenges of dealing with high-dimensional, imbalanced, and non-linear signals, along with complex social and mental health patterns. Classical ML methods, reliant on manual feature engineering, struggle with these challenges. Consequently, DL approaches have become dominant, often outperforming classical techniques without extensive feature design \cite{Zhang2022-gm}. Given the textual nature of much OSM data, recurrent neural networks (RNNs) and their variants, such as LSTMs and GRUs, are widely used to capture sequential patterns, alongside other DNNs for diverse representations. As these models grow more complex, their decision-making becomes increasingly opaque, highlighting the need for XAI methods to reveal which linguistic, temporal, or social cues drive predictions.

\subsubsection{Convolutional representation learning for MDD}
Convolution represents the combination of two signals to produce a third signal, which forms the basis of convolutional neural networks (CNNs) \cite{Lecun1998-pu}. CNNs can recognise patterns in textual embeddings, making them useful for detecting mental disorder signals in OSM posts. For instance, Gaur et al. \cite{Gaur2019-tr} created a suicide risk severity dictionary and a labelled Reddit dataset for multiclass risk detection using CNNs, while Zogan et al. \cite{Zogan2023-by} combined CNNs with hierarchical attention to detect depressive content on $\mathbb{X}$ during the pandemic. CNN-based approaches typically rely solely on text, overlook social interactions, and often produce coarse-grained labels. As CNNs focus on local patterns, XAI methods are essential to reveal which n-grams or feature maps drive predictions.


\subsubsection{Sequential representation learning for MDD} 

Recurrent neural networks (RNNs) excel at modelling long-term sequential data from OSM, outperforming classical MLPs and CNNs due to effective parameter sharing. Textual posts represented as sequences allow RNNs to capture relationships among words and sentences. Vanilla RNNs update hidden states recursively but suffer from vanishing gradients, limiting long-range dependency capture. LSTMs \cite{Hochreiter1997-jc} address this via gated architectures (\textit{input}, \textit{forget}, \textit{output} gates) that maintain long-term context in the cell and hidden states; bidirectional LSTMs (BiLSTMs) further enrich context by processing sequences forward and backward. GRUs \cite{Chung2014-ss} simplify LSTMs using combined \textit{reset} and \textit{update} gates, offering similar performance with fewer parameters, and bidirectional GRUs (BiGRUs) similarly enhance contextual modelling.

\noindent \textbf{MDD using RNNs.} RNNs and their variants are widely applied to MDD for capturing sequential dependencies and temporal dynamics in OSM data \cite{9447025,uban-etal-2021-understanding,thamrin2025distinguishing}. Unlike models treating posts independently, RNNs process sequences of posts or tokens to reveal evolving emotional and linguistic patterns, aiding understanding of depressive symptom progression. Ghosh and Anwar \cite{9447025} framed MDD as a regression task to estimate depression intensity, using an LSTM with Swish activation and weakly labelled $\mathbb{X}$ data via self-supervised learning. However, their approach lacks clinical validation and ignores social interactions and informal language. Thamrin and Chen \cite{thamrin2025distinguishing} combined LSTM sequences with BERT features to capture emotional fluctuations over time, distinguishing depression from bipolar disorder. Despite their effectiveness, RNNs rely on opaque hidden states, making XAI essential to identify which temporal cues (e.g., emotional shifts, linguistic transitions) drive predictions.


\subsubsection{Attention and contextual representation for MDD} 

OSM posts often reveal clear signs of mental health issues, and focusing on these cues is crucial for accurate MDD. Attention mechanisms allow models to assign higher importance to informative words or phrases, improving detection performance \cite{Bahdanau2014-ie,Uban2021-zw}. Transformers \cite{Vaswani2017-mh} replace recurrent models with attention-based architectures that capture long-range dependencies efficiently. Transformer models such as BERT \cite{Devlin2018-zf} and its variants (e.g., DistilBERT \cite{Sanh2019-bi}, RoBERTa \cite{Liu2019-vw}) provide deep contextual representations that can be fine-tuned for MDD on social media data. These models excel at understanding complex language patterns, enabling more accurate and context-aware mental disorder detection.

{\noindent\textbf{MDD using transformers.} Transformer-based models have advanced mental health analytics on social media by leveraging contextualised representations and self-attention to capture subtle linguistic and affective cues \cite{poswiata-perelkiewicz-2022-opi,10.1145/3404835.3462938,10.1145/3543507.3583863,Ragheb2023-nb,Cao2019-fp,Matero2019-st,abuhassan2023classification,wang2025end,liu2024depression,thamrin2025distinguishing}. Poświata et al. \cite{poswiata-perelkiewicz-2022-opi} fine-tuned \textit{RoBERTa} for depression detection (\textit{DepRoBERTa}), revealing fine-grained depression features. Thamrin and Chen \cite{thamrin2025distinguishing} combined BERT embeddings with sequential attention layers to differentiate depression from bipolar disorder, capturing short-term mood fluctuations while maintaining interpretability. Wang et al. \cite{wang2025end} proposed E2-LPS, integrating BERT embeddings with a transformer encoder and a BDI-II-guided post-screening mechanism to focus on clinically relevant posts. Liu et al. \cite{liu2024depression} introduced DeCapsNet, a hierarchical BERT-based capsule network using contrastive learning for richer depression representations. Ragheb et al. \cite{Ragheb2023-nb} developed an ensemble transformer model for detecting at-risk users across multiple disorders by introducing noise into contextual embeddings for better generalisation. Cao et al. \cite{Cao2019-fp} combined masked language models with LSTMs to detect latent signs of suicide ideation. Zogan et al. \cite{10.1145/3404835.3462938} proposed DepressionNet, integrating BERT, BART-based summarisation, BiGRU, and attention layers with user behaviour vectors. Abuhassan et al. \cite{10.1145/3543507.3583863} introduced EDNet, a multimodal transformer framework that fuses textual posts, user biographies, and behavioural patterns for classifying eating disorder-related users.}

Overall, transformer-based models outperform earlier DL approaches for MDD by capturing long-range dependencies, contextual nuance, and affective signals. Their performance improves further when combined with temporal modelling, domain knowledge, or hierarchical reasoning. Multi-head attention offers natural interpretability, highlighting the words, posts, or temporal segments driving predictions. Most methods rely on textual signals, treat disorders as static or binary, and neglect social interactions and network context.

\subsubsection{Graph representation learning for MDD} 

Social connections strongly influence mental health, with individuals significantly more likely to experience depression if their friends do \cite{Rosenquist2011-jx}. This highlights the need to go beyond isolated textual cues and consider users’ relational context. Graph-based representations of social networks capture structural patterns, community interactions, and information flow, providing insights into support systems, isolation, and the spread of mental health signals \cite{Yao2019-vx,Liu2020-ah,Velickovic2017-jd,10.5555/3294771.3294869,Kipf2016-af}. Integrating these network representations with textual content offers a more holistic view of users’ mental wellbeing. Graph neural networks (GNNs), including GCNs \cite{Kipf2016-af}, GATs \cite{Velickovic2017-jd}, GraphSAGE \cite{10.5555/3294771.3294869}, and textual graph embeddings \cite{Yao2019-vx,Liu2020-ah}, extend deep learning to such graph structures, learning from both user interactions and text-derived graphs. By modelling social, semantic, and temporal relationships, these approaches reveal patterns that purely text-based methods may miss, supporting more comprehensive and interpretable MDD.

\noindent\textbf{MDD using GNNs.} Recent studies have adapted GNNs to capture structural patterns in mental health data. They follow two main directions: text-centric graphs \cite{10.1145/3485447.3512128,Naseem2023-xh,abuhassan2025lexicon} and social-network graphs \cite{Pirayesh2021-zm,Sawhney2021-wd,10027637,kuo2023dynamic}. Text-based approaches analyse semantic and temporal patterns in posts. Naseem et al. \cite{10.1145/3485447.3512128} use TextGCN with BiLSTM and attention for four-class depression severity on Reddit. GHAN \cite{Naseem2023-xh} combines TensorGCN embeddings with attentive transformers to detect multi-level suicide ideation. Abuhassan et al. \cite{abuhassan2025lexicon} introduce SEEDNet, merging lexicon-guided GCN/GAT with affective features for eating disorder severity estimation. While effective, these approaches rely solely on post content, missing behavioral, historical, or social cues.  

Social-network-based GNNs model users’ interactions to capture social contagion and network influences. Pirayesh et al. \cite{Pirayesh2021-zm} created PsycheNet dataset from $\mathbb{X}$ and proposed MentalSpot, embedding user tweets with GloVe and 1D convolutional maps across top-k friends. It relies solely on textual data and frames MDD as a binary classification task. Sawhney et al. \cite{Sawhney2021-wd} introduced hyper-SOS for suicide detection. It combines users’ historical posts and social interactions. Textual posts are embedded with BERT, and the HEAT mechanism aggregates posting history. The resulting graph is processed through a Hyperbolic GCN (HGCN) \cite{Chami2019-sc}, which leverages hyperbolic geometry to capture long-
range connections and improve robustness, performance, and interpretability. However, the model uses static graphs. Further, MentalNet \cite{10027637} models users as a heterogeneous ego network over PsycheNet-G, incorporating LSTM autoregressive embeddings, GCN, and convolution layers to capture both tweet content and interactions. Kuo et al. \cite{kuo2023dynamic} extend this to dynamic temporal heterogeneous graphs with ContrastEgo, combining GNNs and a transformer encoder with contrastive learning to model evolving interactions. Despite these innovations, most studies rely on textual posts, perform binary classification, overlook behavioural or user-specific features, and face complexity and explainability challenges, highlighting the need for interpretable models that integrate textual, temporal, and relational factors in line with XAI principles.

\subsubsection{Large language models for MDD}

LLMs are able to capture complex semantic, syntactic, and contextual relationships, and can perform a wide range of NLP tasks with minimal supervision. In mental health, LLMs hold significant potential for early detection, symptom assessment, and severity estimation of disorders. {They can model complex contextual and temporal patterns across user posts, generate human-interpretable rationales, and facilitate multi-role interactions among patients, caregivers, and clinicians \cite{10.1145/3589334.3648137,xu2024mental,zhou2025intermind}. Unlike earlier DL or transformer-only models, LLMs naturally support explanatory reasoning through prompting, self-reflection, and structured outputs, making them well aligned with XAI goals. They can further incorporate domain knowledge through instruction fine-tuning, retrieval-augmented generation (RAG), and chain-of-thought reasoning, enhancing interpretability and alignment with clinical diagnostic standards (e.g., DSM-V).
}

\footnotesize
\begin{table*}[t]
\footnotesize
\caption{Significant ML studies on MDD from online social media data that face challenges with explainability.}
\label{tab:dl_overview1}

\setlength{\tabcolsep}{4pt}

\begin{tabular}{|p{0.6cm}|p{2.2cm}|p{3.9cm}|p{4.3cm}|p{4.5cm}|p{0.7cm}|}
\hline
\multirow{2}{*}{\textbf{Ref}} & 
\multirow{2}{*}{\textbf{Disorder}} & 
\multirow{2}{*}{\textbf{Methodology}} & 
\multicolumn{2}{p{8.4cm}|}{\centering\textbf{Our comments}} & 
\multirow{2}{*}{\textbf{XAI}} \\
\cline{4-5}
& & & \textbf{Pros} & \textbf{Cons} & \\
\hline

\cite{KADKHODA2022101042} & Bipolar Disorder & Behaviour change graph, Stochastic gradient descent, Random forest, KNN, DT, LR, SVM & Mood changes are considered & Non-robust classifier, Unconsidered OSM post contents and social network. Focuses only on disorder existence & $\times$ \\
    \hline     
    \cite{Sawhney2021-wd} & Suicide Ideation & BERT, Hawkes, Hyperbolic GCN  & User interaction, Contextualised embedding & Static social network, only post content based node features. Focuses only on disorder existence & $\times$ \\
    \hline
    \cite{10.1145/3543507.3583863} & Eating Disorders & BERT, TCN, BiGRU  & User-type identification, Multi-modality, Contextual embedding & No social network & $\times$ \\
    \hline
    \cite{10.1145/3404835.3462938} & Depression & BERT, KMeans, BART, BiGRU, CNN  & Multi-modality, Contextual embedding & No social network, Focuses only on disorder existence & $\times$ \\
    \hline
    \cite{10027637}  & Depression & GCN, BiLSTM, CNN  & Augmented dataset, User interaction & Static social network, Only post contents as node features, Focuses only on disorder existence & $\times$ \\
    \hline
    \cite{kuo2023dynamic} & Depression & RoBERTa, GCN, Transformer & Dynamic social network, User interactions, Contextual embedding & Only post contents as node features, Focuses only on disorder existence & $\times$ \\
    \hline
    \cite{Sawhney2020-en} & Suicide Ideation & BERT, LSTM, Transformer & Contextual Embedding, Time-aware detection, Considers historical posts & No social network, Only post contents, Focuses only on disorder existence  & $\times$ \\
    \hline
    \cite{9308975} & Suicide Ideation & Knowledge graph, BERT, CNN, LSTM & Graphical representation, Contextual embedding, Multi-modal data & No social network, Focuses only on disorder existence & $\times$ \\
    \hline
    \cite{Ragheb2023-nb} & Depression, Anorexia, Self-harm and Suicide & Transformer-based encoders, Negatively correlated noisy learners & Contextual embedding, Noisy learners, Multiple disorders & No social network, Only post contents, Focuses only on disorder existence & $\times$ \\
    \hline 

    \cite{ZHANG2024111650} & Depression & BERTopic, MLP, GNN, Temporal LSTM, Attention, PLM & Contextual Embedding, Historical user posts, Graphical representation & No social network, Focuses only on disorder existence, Relies only on textual data. & $\times$   \\ 
    \hline  
    \cite{10241281} & Depression & LDA, ResNet50, OCR, Ensemble Learning & Multi-modal features, Ensemble of classifiers, User and depression specific features & Focuses only on disorder existence, Non-contextual embedding, No social network & $\times$   \\ 
    \hline 
\end{tabular}
\end{table*}
\normalsize

\footnotesize
\begin{table*}[htbp]
\footnotesize
\caption{Significant ML studies on MDD from online social media data that do not explicitly focus on explainability but have potential for explanation.}
\label{tab:dl_overview2}

\setlength{\tabcolsep}{4pt}

\begin{tabular}{|p{0.6cm}|p{2.2cm}|p{3.8cm}|p{4.4cm}|p{4.5cm}|p{0.7cm}|}
\hline
\multirow{2}{*}{\textbf{Ref}} & 
\multirow{2}{*}{\textbf{Disorder}} & 
\multirow{2}{*}{\textbf{Methodology}} & 
\multicolumn{2}{p{8.4cm}|}{\centering\textbf{Our comments}} & 
\multirow{2}{*}{\textbf{XAI}} \\
\cline{4-5}
& & & \textbf{Pros} & \textbf{Cons} & \\
\hline
    \cite{Cao2019-fp} & Suicide Ideation & LSTM, Attention, ResNet & Extracts latent information, Contextual embedding, Multi-modal data & No social network, Focuses only on disorder existence  & $\checkmark$ \\
    \hline 
    \cite{9733425} & Depression & LSTM, Attention, LR,  Ensemble Methods & Sentiment Lexicons, symbolic and subsymbolic models & Only text based features, Focuses only on disorder existence, Non-contextual embeddings & $\checkmark$  \\
    \hline 
    \cite{10.1145/3437963.3441805} & Suicide Ideation & Longformer, BiLSTM, Attention & Ordinal classification, Contextualised embedding & No social network & $\checkmark$ \\
    \hline
    \cite{10.1145/3485447.3512128} & Depression & TextGCN, BiLSTM, Attention  & Ordinal classification, Contextualised embedding & No social network, Only post contents & $\checkmark$ \\
    \hline
    \cite{Zogan2023-by}  & Depression & CNN, MLP, Attention & Considers both word and sentence importance & No social network, Only post contents, Focuses only on disorder existence & $\checkmark$\\
    \hline
    \cite{zhang-etal-2023-sentiment} & Depression & Mental RoBERTa, Transformer, Multi-head soft attention  & Ordinal classification, Contextual embedding & No social network, Only post contents & $\checkmark$ \\
    \hline
    \cite{9349170} & Suicide Ideation & LIWC, LSTM, Attention & Considers multi-sources, Comprehensive textual analysis & No social network, Only post contents, No contextual embedding, Focuses only on disorder existence & $\checkmark$ \\
    \hline
    \cite{10.1145/3543507.3583867} & Depression & BERT, Transformer, Self attention & Real time detection, Contextual mood and content embedding & No social network, Only post contents, Focuses only on disorder existence & $\checkmark$ \\
    \hline
    \cite{sawhney-etal-2022-risk} & Suicide Ideation & BERT, Bi-LSTM, MLP, Attention & Contextual embedding, Ordinal classification, Robust model & No social network, Only post contents & $\checkmark$ \\ 
    \hline
    \cite{wang-etal-2021-learning} & Suicide Ideation & Doc2Vec, Multi-head attention, & Extracts emotions, Detects disorder couple of months earlier & No social network, Only post contents, No contextual embedding, Focuses only on disorder existence & $\checkmark$ \\
    \hline
    \cite{abuhassan2025lexicon} & Eating Disorders & BERT, Lexicon-guided GNN, GCN, GAT & Fine-grained severity estimation, Integrates affective features, Lexicon-based graph attention improves interpretability & Only ED-related content, Requires domain-specific lexicon & $\checkmark$ \\
    \hline
    \cite{poswiata-perelkiewicz-2022-opi} & {Depression} & RoBERTa & Contextual Embedding & No social Network, Only post contents & $\checkmark$ \\ 
    \hline
    \cite{yoon2022d} & {Depression} & Transformers & Contextual embedding, Multi-modality & No social network & $\checkmark$ \\
    \hline
    \cite{min2023detecting} & {Depression} & OpenSmile\footnote{\url{https://www.audeering.com/research/opensmile/}}, eGeMAPS, CNN,  XGBoost & Multi-modality & No social network & $\checkmark$ \\
    \hline
\end{tabular}
\end{table*}
\normalsize

\footnotesize
\begin{table*}[htbp]
\footnotesize
\caption{Significant ML studies on MDD from online social media data that focus on explainability.}
\label{tab:dl_overview3}

\setlength{\tabcolsep}{4pt}

\begin{tabular}{|p{0.6cm}|p{2.3cm}|p{3.8cm}|p{4.4cm}|p{4.4cm}|p{0.7cm}|}
\hline
\multirow{2}{*}{\textbf{Ref}} & 
\multirow{2}{*}{\textbf{Disorder}} & 
\multirow{2}{*}{\textbf{Methodology}} & 
\multicolumn{2}{p{8.4cm}|}{\centering\textbf{Our comments}} & 
\multirow{2}{*}{\textbf{XAI}} \\
\cline{4-5}
& & & \textbf{Pros} & \textbf{Cons} & \\
\hline
    \cite{alghazzawi2025explainable} & {Suicide Ideation} & SVM, LR, GB, DT, RF & Transparent analysis & Uncontextual embedding, No social network, Only post content & $\checkmark\checkmark$ \\
    \hline    
    \cite{hameed2025explainable} & {Depression} & TF-IDF, N-grams, BoW, LDA, GloVe, SVM, RF, XGB and ANN & High Accuracy, Computational Inexpensivity & No social network, Only post contents, Focuses only on the existence of the disorder &  $\checkmark\checkmark$ \\
    \hline
    \cite{al2025effective} & {Depression} & RF, SVM, MLP, CNN, BERT, DistilBERT, RoBERTa & Contextual Embedding, Multi modal approach & No social network & $\checkmark\checkmark$ \\
    \hline
    \cite{Amini2020-zk} & Eating Disorder (Anorexia Nervosa) & ELMo, CNN, Attention & Contextual embedding & No social network, Only post contents, Focuses only on disorder existence & $\checkmark\checkmark$\\ 
    \hline
    \cite{han-etal-2022-hierarchical} & Depression & BERT, HAN & Adopts metaphor generation, Contextual embedding & No social network, Only post contents, Focuses only on disorder existence & $\checkmark\checkmark$ \\
    \hline
    \cite{Zogan2022-wd} & Depression & GloVe, MLP, HAN & Multi-modality & No social network, No contextual embedding, Focuses only on disorder existence & $\checkmark\checkmark$ \\
    \hline
    \cite{Naseem2022-vt} & Suicide Ideation & TensorGCN, Longformer, Transformer & Ordinal classification, Contextualised embedding, Performs well on long sequences & No social network, Only post contents & $\checkmark\checkmark$\\
    \hline
    \cite{ibrahimov2025depressionx} & Depression & Text + Depression-specific KG, Residual attention & Multi-level attention, Knowledge-infused, Fine-grained severity estimation, Clinically interpretable & Requires knowledge graph, Limited generalisation outside annotated concepts & $\checkmark\checkmark$ \\
    \hline    
    \cite{belcastro2025detecting} & {Depression} & BERTweet, LSTM, ChatGPT & High Accuracy, Contextual Embedding & No social network, Only post contents & $\checkmark\checkmark$ \\
    \hline   
    \cite{Bao2024} & {Depression} &  WT5, WBART, BERT, T5 & Symptom based analysis, Contextual embedding, Clinically grounded explanations & No social network, Only post contents & $\checkmark\checkmark$ \\
    \hline  
    \cite{pmlr-v284-mahdavinejad25a} & {Depression} & Sentence Encoder, GPT-4, Symptom Prototype & Contextual Embedding, Symptom-aware analysis &  No social network, Only post contents  & $\checkmark\checkmark$ \\
    \hline
\end{tabular}
\end{table*}
\normalsize

MentaLLaMA \cite{10.1145/3589334.3648137}, Mental-LLM \cite{xu2024mental}, and InterMind \cite{zhou2025intermind} represent recent LLM-driven approaches to MDD.
MentaLLaMA is an open-source instruction-tuned variant of LLaMA2 developed specifically for explainable depression detection. The authors built IMHI, a large multi-task dataset containing 105K OSM posts covering binary and multi-class MDD, cause identification, and wellness factor analysis. They also used few-shot prompting with ChatGPT to generate rationales, enabling the model to provide interpretable outputs. Several versions of MentaLLaMA (7B, 7B-chat, 13B-chat) were fine-tuned on this dataset to evaluate the effect of model scale on both prediction and explanation quality. Similarly, Mental-LLM \cite{xu2024mental} offers a broader benchmark across multiple disorders (depression, anxiety, and stress), evaluating a range of LLMs such as Alpaca, FLAN-T5, LLaMA2, GPT-3.5, and GPT-4. The study compares zero-shot, few-shot, and instruction-tuned settings, showing that smaller domain-adapted models (e.g., Mental-Alpaca, Mental-FLAN-T5) can outperform larger general-purpose LLMs in balanced accuracy. Their analysis of reasoning traces illustrates that LLM explanations are promising but still inconsistent, suggesting that unstructured chain-of-thought is not yet fully reliable for clinical interpretation. InterMind \cite{zhou2025intermind} moves beyond single-turn prediction by introducing an interactive framework designed for doctor–patient–family assessment. It includes an AI Psychological Chatbot that gathers emotional and behavioural information through conversational prompts, and an AI Psychiatrist that produces structured diagnostic reports covering depression severity, symptoms, and treatment suggestions. The system integrates prompt engineering, chain-of-thought reasoning, and RAG to support more stable explanations and reduce hallucinations, while ensuring closer alignment with DSM-V criteria. This illustrates a shift from static classifiers to conversational and clinically grounded LLM systems capable of structured, explainable decision support. 

In summary, ML models in Tables~\ref{tab:dl_overview1}–\ref{tab:dl_overview3} achieve promising MDD results. However, high-accuracy datasets often rely on few anchor posts explicitly mentioning conditions (e.g., ``I was diagnosed with depression''), limiting detection of implicit or early-stage symptoms. Incorporating historical posts and user interactions can reveal latent signals, yet such models are complex and less explainable. LLM-based pipelines integrate reasoning and knowledge but remain opaque, highlighting the need for XAI-oriented designs (Section \ref{sec:explainability}).

\subsection{Importance of domain knowledge}

Domain knowledge grounds AI models in clinically meaningful concepts. It enhances interpretability, stabilises predictions, and aligns outputs with established mental-health understanding.


\subsubsection{Structured resources} 
Knowledge graphs provide explicit, interpretable representations of clinical and biomedical relationships. Gao et al. \cite{gao2025large} introduced the Mental Disorders Knowledge Graph (MDKG), integrating over 10 million relationships among 1.6 million entities from literature and databases (DrugBank, DisGeNET, UMLS) using GPT-4 reasoning, BERT-based entity extraction, and active learning. Its contextual attributes strengthen traceability and interpretability, while achieving 79\% correctness and a 70\% usability improvement. Ibrahimov et al. \cite{ibrahimov2025depressionx} built \textit{DepressionX}, a knowledge-infused attention model for explainable depression severity estimation. It fuses textual embeddings with a depression knowledge graph via multi-level attention and graph reasoning for clinically interpretable outputs.


\subsubsection{Observational studies} 

Observational and psycholinguistic studies inform model design by revealing ground-truth behavioural patterns. Nguyen et al. \cite{Nguyen2014-pq} analysed LiveJournal users’ mood-tagged posts, extracting sentiment via ANEW \cite{Bradley1999AffectiveNF}, psycholinguistic features via LIWC, and topics via LDA. Differences between clinical and control groups revealed more frequent negative emotions among clinical cases. Suicide-related words (e.g., coffin, kill, bury) also appeared more often in higher-risk groups, forming interpretable linguistic markers that continue to inform model explanations in later work.


\subsubsection{Computational approaches} 

To overcome social stigma and limited clinical assessments, computational methods approximate validated tools. These approaches often enhance explainability by grounding predictions in clinically validated structures such as questionnaires or symptom sets. Perez et al. \cite{PEREZ2022102380} estimated depression intensity of an OSM user by automatically responding to the BDI-II questionnaire (See Section \ref{sec:traditionalmethods}) on users' behalf. The authors leveraged eRISK-2019 \cite{Losada2019-cm}  and eRISK-2020 \cite{10.1007/978-3-030-45442-5_72} datasets, which contain Reddit posts as well as answers to the BDI-II questionnaire for a sample of users. To enhance clinical relevance, they later developed a sentence dataset, BDI-Sen \cite{10.1145/3539618.3591905}, which is specifically designed to capture clinical symptoms associated with depression. The dataset, derived from eRISK-2019, uses sentence embeddings and similarity measures to identify content relevant to symptoms as assessed by the BDI-II questionnaire, providing fine-grained, interpretable supervision signals. Similarly, Anwar et al. \cite{9906413} created EDBase, a lexicon of ED terminology and relevance scores, enabling structured, interpretable feature extraction for deep learning models \cite{10.1145/3543507.3583863,abuhassan2025lexicon}.

\subsubsection{Domain-aligned embeddings}
Domain-aligned embeddings are crucial for capturing nuanced patterns in mental health text. Deep embedding models trained on mental health corpora provide rich representations that improve downstream analysis while also supporting more transparent and clinically grounded interpretations. For example, \textit{MentalBERT} \cite{Ji2021-hu} is a BERT-based model pre-trained on Reddit posts related to mental health, offering robust embeddings for MDD in OSM text. By embedding domain-relevant expressions, symptoms, and linguistic markers, these models help explanation methods (e.g., attention, feature attribution) focus on clinically meaningful cues rather than generic linguistic patterns, thereby strengthening the reliability of interpretations.

\subsection{Explainability is what we need}
\label{sec:explainability}

Explainability provides insights into how AI models reach their predictions, complementing domain knowledge and supporting informed decision-making. This section reviews how the MDD models in Section~\ref{sec:mlmdd} incorporate XAI approaches. It covers interpretable models, post-hoc methods, attention mechanisms, graph-based explanations, and emerging LLM-based techniques, while highlighting their strengths and limitations in MDD. Table~\ref{tab:multicol} summarises the main explainability categories, representative examples, and their applications.

\begin{table*}[!ht]
\caption{Overview of Explainability methods}
\begin{tabular}{|p{1.5cm}|p{2.6cm}|p{1.2cm}|p{2.0cm}|p{5.6cm}|p{2.5cm}|}
    \hline
    \multicolumn{2}{|c|}{\textbf{Explanation Type}} & \textbf{Scope} & \textbf{Model Type} & \textbf{Example}& \textbf{Applied for MDD?}\\
    \hline
    \multicolumn{2}{|c|}{Interpretable Models} & Global & Model Specific & Linear and Logistic Regression, Decision Trees & $\checkmark$ \\
    \hline
    \multirow{9}{4em}{Post-hoc}&\multirow{3}{*}{Feature relevance}&Local&Model Agnostic& SHAP, KernelSHAP & $\times$\\
    \cline{3-6} 
    &  &Local&Model Specific&
    Layer-wise Relevance Propagation (LRP) & $\times$\\
    \cline{3-6} 
    
    &  &Local&Model Specific&A-Grad and RePAGrad & $\times$\\
    \cline{2-6} 
    & \multirow{2}{*}{By Approximation} &Local&Model Agnostic&LIME & $\checkmark$ \\ 
    \cline{3-6} 
    &  &Global&Model Agnostic&PGExplainer (for graphs) & $\times$\\
    \cline{2-6} 
    & \multirow{2}{*}{By Example} &Local&Model Specific&GNNExplainer and GCN-SE (for graphs) & $\times$ \\
    \cline{3-6} 
    &  &Local&Model Specific&Attention & $\checkmark$\\
    \cline{2-6} 
    & \multirow{2}{*}{By Vis. Explanation} & Local& Model Agnostic & Individual Conditional Expectation (ICE) & $\times$ \\
    \cline{3-6} 
    &  & Global &Model Agnostic & Partial Dependence Plot (PDP) & $\times$\\
    \hline
\end{tabular}
\label{tab:multicol}
\end{table*}

\subsubsection{Interpretable models}

Some models, such as linear regression, logistic regression, and decision trees, stand out for their inherent interpretability. Their internal architectures make it easy to understand the rationale behind the models' decisions. Linear Regression and Logistic Regression provide transparency through the generation of feature weights. These weights assign significance to individual features, enabling us to recognise their importance in influencing predictions. Decision Trees, on the other hand, adopt a hierarchical structure of questions, forming a tree-like visualisation. This representation offers an intuitive way to grasp the decision logic. By following the branches of the tree, we can trace the series of questions and conditions that lead to a particular outcome. Inherently interpretable models have been applied in various mental health contexts. For instance, Kelly et al. \cite{kelly2025interpretable} proposed an interpretable model for predicting mental health treatment outcomes, combining probabilistic methods with a self-interpretable logistic regression approach. Their dataset included patient responses from standardized questionnaires, such as the PHQ-9 (depression), GAD-7 (anxiety), SIAS (social anxiety), PDSS (panic disorder), and the Fear Questionnaire, along with demographic and medical history data. Similarly, Liu et al.  \cite{LIU2025529} developed an interpretable machine learning model to screen for depression in functionally disabled older adults in China. They analysed data from the 2020 CHARLS (China Health and Retirement Longitudinal Study) database, identifying key depression predictors using LASSO regression and logistic regression techniques. While both studies contribute to explainable mental health analysis, their approaches have limitations. First, the required data (clinical questionnaires, medical records) are not as easily accessible as public sources like OSM. Second, these methods rely solely on statistical correlations, ignoring semantic and sentiment-based insights. As a result, their explanations highlight feature importance statistically rather than capturing deeper mental health context. Finally, the models' simplicity restricts their ability to analyse complex, high-dimensional features.


\subsubsection{LIME and SHAP} \label{sec:lime}

Some commonly used post-hoc explainability methods include LIME \cite{Ribeiro2016-cn} and SHAP \cite{Lundberg2017-xf}. While both LIME and SHAP are model-agnostic, LIME approximates model behaviour locally, whereas SHAP distributes contribution scores fairly across all features using cooperative game theory.

LIME \cite{Ribeiro2016-cn} is an acronym for local interpretable model-agnostic explanations. 
It is a widely-used model-agnostic method for providing local explanations. Its primary aim is to explain complex ML models by approximating their behavior with a simpler and more interpretable model, such as linear regression or logistic regression. The local neighborhood is created by generating a dataset through a perturbation method, which introduces controlled noise to the input data. Mathematically, given a prediction function $f: \mathbb{R}^n \to \mathbb{R}$ and a data point $x$, LIME seeks a simpler model $g(x)$ within a local neighborhood of $x$. This is done by minimising the loss between the original complex model and the simpler model using Equation \ref{eqn:lime}, where $\mathcal{L}(f, g, \pi_x)$ represents the loss function shown in Equation \ref{eqn:limeloss}, and $\Omega(g)$ is a regularization term applied to the simpler model $g$. The goal is to find the parameters for this simpler model that best approximate the behavior of the complex model within the local neighborhood. In Equation \ref{eqn:limeloss}, $\mathcal{Z}$ denotes the set of points within the local neighborhood of $x$, $\pi_x(z)$ denotes the weight assigned to data point $z\in \mathcal{Z}$ determined using a perturbation method, and $f(z)$ and $g(z')$ are the predictions made by the complex and simpler models, respectively. 
\begin{eqnarray} 
\xi(x) = \argmin_{g \in G} \left[\mathcal{L}(f, g, \pi_x) + \Omega(g)\right] \label{eqn:lime}\\
\mathcal{L}(f,g,\pi_x) = \sum_{z,z' \in \mathcal{Z}} \pi_x(z) \cdot \left(f(z) - g(z')\right)^2 \label{eqn:limeloss}
\end{eqnarray}
LIME has proven valuable in various applications, including causal analysis of depression cases \cite{Saxena2022-xp}. The study used LIME to compare model-generated explanations with human-generated explanations.

SHAP \cite{Lundberg2017-xf}, an acronym for SHapley Additive exPlanations, can be used to obtain the importance (called Shapley value) of individual features used in predictive models. The Shapley value  $\phi_i(f, x)$ of a specific feature $i$ of a data point $x$ for a complex model $f$ is calculated using Equation \ref{eqn:shapely}, where $x'$ denotes the simplified input set (often a subset of the complete set of features), $M$ is the total number of available features, $|z|!$ calculates the factorial of the size of subset $z$, $f_x(z')$ denotes the model's prediction for $z'$ subset given as input, and $f_x(z' \backslash i)$ denotes the model's prediction after removing feature $i$ from the subset $z'$. 

\relsize{-1}
\begin{equation} \label{eqn:shapely}
\phi_i(f, x) = \sum_{z' \subseteq x'} \frac{|z|!(M - |z'| - 1)!}{M!} \left[ f_x(z') - f_x(z' \backslash i) \right]
\end{equation}
\normalsize
Shapley values tend to perform well with linear models, but may encounter limitations with complex models. An extension of SHAP, known as KernelSHAP, is tailored to handle complex models by breaking down Shapley values into smaller subsets of features and employing Monte Carlo sampling techniques to approximate these values.

LIME and SHAP based post-hoc explainability methods have been frequently applied in mental health analysis. Kerz et al. \cite{kerz2023toward}, Saxena et al. \cite{Saxena2022-xp}, Hameed et al. \cite{hameed2025explainable}, Masud et al. \cite{al2025effective}, Lamba et al. \cite{Lamba2026} and Alghawazzi et al. \cite{alghazzawi2025explainable} developed ML models to analyse mental disorders using OSM data, employing LIME and SHAP to explain their predictions. Alghawazzi et al. \cite{alghazzawi2025explainable}  focused on explainable detection of suicidal ideation using an ensemble learning approach. They applied TF-IDF vectorization to extract text features and used LIME to highlight important words in their predictions. Kerz et al. \cite{kerz2023toward} studied mental health challenges, including ADHD, anxiety, bipolar disorder, depression, and stress, by analysing linguistic and lexical features. These features were then processed using a Bidirectional LSTM model, with LIME and SHAP providing explanations. However, both  Kerz et al. \cite{kerz2023toward} and Alghawazzi et al. \cite{alghazzawi2025explainable} rely on statistical existence of features, meaning LIME and SHAP only highlight important statistical features without capturing deeper semantic or sentiment-based insights. Saxena et al. \cite{Saxena2022-xp}  also used a Bidirectional LSTM model, but with pre-trained FastText word embeddings, to predict mental disorder causes from OSM text. While LIME identifies key tokens in individual posts, its explanations remain limited. It provides local, post-specific approximations rather than a comprehensive understanding of the model’s decision-making process.

\subsubsection{Explanation by Attention values} \label{sec:explainabilitybyattention}

Attention mechanisms \cite{Bahdanau2014-ie, Vaswani2017-mh} are widely used in NLP to provide word-level importance scores, which offers an intuitive form of model explainability \cite{wiegreffe2019attention}. Some studies \cite{Naseem2022-vt, Amini2020-zk} have leveraged attention to detect suicide risk from Reddit data, where high-weight words highlight indicators of suicidal ideation. Beyond word-level explanation, hierarchical attention networks (HAN) \cite{Yang2016-xr} extend this approach to the sentence level. Word-level attention values are first computed, followed by sentence-level attention values. For example, MDHAN \cite{Zogan2022-wd} detects depressed users while providing model explanations using HAN. It encodes user posts at both the tweet and word levels, offering explanations at multiple granularities. While MDHAN is effective for both classification and explainability, it focuses solely on textual features and does not consider user behavior patterns. Similarly, \cite{han-etal-2022-hierarchical} applied hierarchical attention mechanisms to explain depression detection on Twitter. However, their approach incorporates metaphor mapping, but the role of metaphors in explainability is not well justified. 

A known limitation of attention-based explanations is that they do not indicate whether a highlighted word or tweet contributes positively or negatively to a model's decision. To address this, \cite{9671639} introduced the Attention Gradient (AGrad) mechanism, which quantifies the directional contribution of contextual embeddings. Building on this idea, \cite{kerz2023toward} applied AGrad to analyse the impact of linguistic and lexical features in the study of mental disorders. DepressionX \cite{ibrahimov2025depressionx} employs multi-level residual attention to provide explanations at the word, sentence, and post levels. Although it integrates external knowledge sources, its explainability primarily arises from attention distributions. Attention can also capture temporal dynamics. Thamrin and Chen \cite{thamrin2025distinguishing} developed an attention-based temporal framework to distinguish depression from bipolar disorder. Their model applies attention weights over grouped time intervals of posts, enabling the identification of emotionally salient periods and fluctuations that influence classification. Attention mechanisms have also been used to interpret the inner workings of transformer-based black-box models such as BERT and GPT \cite{hoover2019exbert, Van_Aken2020-mc, Vig2019-hx}, providing an additional lens for model interpretability. 

However, the effectiveness of attention as an explanation remains debatable. Attention values can be unstable, sensitive to outliers, and may reflect model focus rather than causal influence \cite{hu2023seat, jain2019attention, wiegreffe2019attention}. Consequently, while attention-based methods offer multi-level interpretability in mental health prediction, their outputs should be interpreted cautiously and ideally complemented with other explanation techniques to ensure faithful and robust explanations.

\subsubsection{LRP, PDP, and ICE}
LRP (or layer-wise relevance propagation) \cite{Bach2015-vo} is an explainability method used to analyse the feature contributions of complex models. It works by propagating relevance scores backward through the layers of a neural network model, determining the importance of each neuron all the way back to the input features. 
This process allows us to trace back and understand the contribution of each neuron and input feature in the model's decision-making.



PDP (partial dependence plot) visualises the marginal effect of model features using a partial dependence method \cite{Friedman2001-yz}. Based on this, Greenwell et al. \cite{Greenwell2018-yl} introduced \textit{importance measure}  to compute the importance of a feature. 
While PDP illustrates the average effect of a feature, it does not focus on the prediction changes based on individual instances. This limitation is addressed by ICE (individual conditional expectation) plots \cite{Goldstein2015-fw}. They depict one line for each instance and each feature by manipulating the feature of interest while keeping the other features fixed.

Although LRP, PDP, and ICE are not commonly used in the analysis of mental health challenges and OSM contexts, they still have potential for explaining these analytical models. They may be especially useful for analysing the effects of social, behavioral, emotional, and sentimental features, which could provide insights into model decisions.

\subsubsection{Explainability of GNN-based methods} 

While GNNs achieve strong performance in mental health analytics, making their predictions interpretable, particularly for MDD, remains an open challenge due to their structural complexity \cite{Yuan2022-lk}. Traditional model-agnostic methods such as LIME, attention mechanisms, and LRP have been applied to graph data, but several GNN-specific approaches have also been developed to better reflect the unique properties of graph architectures. Key methods include GNNExplainer \cite{Ying2019-zg}, PGExplainer \cite{Luo2020-tq}, and GCN-SE \cite{Fan2021-bm} (see more details below). A recent method, DepressionX \cite{ibrahimov2025depressionx}, integrates GIN- and GAT-based representations of a depression-specific knowledge graph with textual embeddings from OSM posts. It offers interpretable insights into depression severity by extracting salient subgraphs and highlighting influential depression-related concepts and relations. The model also provides attention-based explanations for the textual component, thereby combining linguistic and graph-based signals into a unified explanation.


GNNExplainer \cite{Ying2019-zg} is a powerful model-agnostic method for explaining GNN predictions by identifying the most influential subgraphs and node features. It can generate explanations for individual nodes, links, or whole graphs, making it useful for understanding why a GNN produces certain outputs. However, it primarily provides local explanations, meaning each instance is interpreted independently and the insights do not easily generalise across the model. Additionally, generating explanations requires retraining for each instance, which can be time-consuming for large graphs. Finally, because it focuses on individual cases rather than the full network, it does not capture a global perspective of the model, limiting its ability to reveal overarching patterns in graph reasoning. Despite these limitations, GNNExplainer remains a valuable tool for interpretable graph-based modelling.
PGExplainer \cite{Luo2020-tq} addresses the limitations of GNNExplainer by providing collective and inductive explanations for GNN predictions. Like GNNExplainer, it maximises the mutual information between the input graph and the subgraph, but it covers multiple instances to offer a more global understanding. Using a parameterised generative model, PGExplainer uncovers structural patterns crucial for predictions, with shared parameters enabling explanations for unexplained nodes without retraining. Both methods work for static graphs, but neither can explain dynamic GNNs.
%
%
GCN-SE \cite{Fan2021-bm} addresses the explainability in dynamic GNNs with the help of attention values. It treats graph snapshots at different times of dynamic graphs as different channels of data and attaches a set of learnable attention weights with them based on GCN \cite{Kipf2016-af} and SE-Net \cite{hu2018squeeze}.
Despite these advances, graph-based explainability has seen limited use in MDD. Given the strong influence of social connections on mental health, GNN-based explainability could offer valuable insights into MDD by jointly analysing user interactions and content embeddings. Moreover, dynamic approaches such as GCN-SE can capture the temporal progression of mental health states. 
\
{
\subsubsection{LLM-based explanations} \label{sec:llmexplanation}

LLMs have emerged as promising tools for generating natural-language explanations of AI predictions. Their ability to produce human-readable rationales makes them appealing for mental health applications. However, the internal mechanisms of LLMs remain largely opaque, posing challenges for trustworthy explainability in MDD analysis \cite{Yang2023-yx}. Several domain-specific LLMs have been developed to support explainable mental health analysis. For example, MentaLLaMA \cite{10.1145/3589334.3648137} generates natural-language rationales describing potential contributors to depression and stress. Similarly, Mental-LLM \cite{xu2024mental} produces structured reasoning aligned with psychological constructs. LLMs have also been used to verbalise explanations of existing predictive models. Belcastro et al. \cite{belcastro2025detecting} employed ChatGPT to explain the outputs of a BERTweet-based classifier by generating textual rationales grounded in model predictions and highlighted input tokens. Likewise, Bao et al. \cite{Bao2024} proposed explanation pipelines that utilise autoregressive transformer models such as WT5 and MentalMistral to generate reasoning-based explanations for mental health predictions.

Despite these advances, most LLM-generated explanations remain \textit{post-hoc}. They provide plausible textual justifications but do not necessarily reveal how predictions are internally formed, meaning the underlying decision-making process remains difficult to interpret. Current research therefore explores techniques to analyse LLM behaviour. Prompt-based probing evaluates model responses to structured instructions to infer encoded knowledge \cite{ju2024large,bang2023multitask,deshpande2023toxicity,yin2023large,xie2023adaptive,wang2023resolving}, while attention analysis and representation probing examine which internal layers capture relevant linguistic or psychological signals \cite{ju2024large}. However, these approaches typically reveal correlations rather than causal reasoning mechanisms.

Emerging research on \textit{mechanistic interpretability} (MI) seeks deeper insights by reverse-engineering the internal computations of LLMs to identify interpretable circuits \cite{bereska2024mechanistic}. For example, Wang et al. \cite{wang2022interpretability} traced attention circuits in GPT-2 responsible for indirect object identification, demonstrating how internal components can correspond to specific reasoning functions. Although promising, MI has not yet been widely applied in mental health applications, where it could potentially uncover clinically meaningful linguistic patterns associated with depressive symptoms. Another promising direction is RAG, which enhances explainability by grounding model outputs in external knowledge sources. Retrieved documents act as explicit evidence supporting generated explanations, improving transparency and reducing hallucinations \cite{ge2025survey,yang2025retrieval,klesel2025retrieval}. In mental health applications, domain resources such as DSM-V diagnostic criteria could be incorporated into the retrieval pipeline to produce explanations grounded in clinically relevant knowledge \cite{kharitonova2025incorporating}.

While LLMs provide powerful mechanisms for generating natural-language explanations, current approaches largely approximate model behaviour rather than exposing its internal reasoning. Advancing explainable LLMs for mental health applications will therefore require deeper interpretability methods and stronger grounding mechanisms to ensure reliable and clinically meaningful explanations.
}

\subsubsection{{Counterfactual Explanations}} \label{sec:counterfactuals}
{Although several post-hoc explanations such as LIME and SHAP are influential to reveal the important input features, they fail to provide the alternative features that can change the model output, by providing counterfactuals \cite{mothilal2020explaining}. Moreover, counterfactual reasonings allows the healthcare providers to investigate the effect of the small input changes on the model decision to examine the model behaviour and potential biases in clinical settings \cite{ali2025artificial}.  In medical settings, counterfactual reasoning yielded significantly better outcomes regarding emotions, explanations and compared to traditional explanation methods such as LIME and LRP \cite{mertes2022ganterfactual}. Moreover, counterfactual reasoning can disclose the mental disorder based symptoms and reinforce the interpretability of AI-based clinical decision support systems.  \cite{qin2025explainable}}

\section{Datasets and Evaluation Methods}
\label{sec:experimentaldesign}

\subsection{Experimental datasets}
Datasets serve as fundamental components for AI- and data-driven decision-making systems. The dataset curated by Shen et al.~\cite{Shen2017-dy} from $\mathbb{X}$ stands out as a frequently utilised resource in research on depression detection. Another notable data source is the eRISK series~\cite{Losada2016-rd}, collected from Reddit, extensively used by researchers for detecting various mental disorders, including depression, EDs, self-harm, and pathological gambling. Dreaddit, compiled by Turcan et al.~\cite{Turcan2019-td}, is a Reddit-based dataset designed for binary depression classification, later enhanced by Naseem et al.~\cite{10.1145/3485447.3512128} for multi-class severity classification. PsycheNet~\cite{Pirayesh2021-zm} is a social-contagion-driven dataset constructed from $\mathbb{X}$, primarily for depression detection. Mihov et al.~\cite{10027637} improved this dataset, creating PsycheNet-G, by incorporating additional features like bidirectional replies, mentions, and quote-tweets to enhance the robustness of social network data. A more recent dataset by Abuhassan et al.~\cite{abuhassan2025lexicon} provides a 6,500-tweet benchmark for estimating eating disorder severity on OSM. Each tweet was annotated under the guidance of clinical experts into one of four severity levels (none, mild, moderate, high). The dataset includes both balanced and imbalanced versions, supporting fine-grained severity modelling beyond binary ED detection. Further details about these and other relevant datasets are available in Table~\ref{tab:experimental_datasets}. {An extensive list of various related datasets can be found here\footnote{\url{https://github.com/bucuram/depression-datasets-nlp?tab=readme-ov-file}}}.

Data collected from OSM for MDD are sometimes annotated using automatic or semi-supervised methods. However, reliable and unbiased diagnostics require annotation by domain experts. Many datasets include anchor posts\footnote{An anchor post is a post in which a user explicitly indicates having a disorder, based on which the user is labelled. For example, ``I have been diagnosed with depression''.} containing explicit signals of mental disorders, which can lead to potential data leakage. To mitigate this issue, there is a need for datasets that include detailed historical user information, such as posts (excluding anchor posts), online activities, and social network interactions. Such data are important not only for robust model training but also for faithful explainability evaluations.
{We also observe a strong reliance on Reddit and $\mathbb{X}$ in the literature. This is largely a consequence of data accessibility, as many other OSM platforms do not provide APIs that enable large-scale data collection for academic research. Moreover, Reddit users often produce longer textual content, which can provide richer signals regarding users’ mental states. Similarly, the dominance of English-language datasets reflects both the prevalence of English in academic research and the demographics of widely studied OSM platforms.}

\begin{table*}[t]
\footnotesize
\caption{Notable experimental datasets ($\checkmark$: publicly available, $\odot$: available on request).}
\label{tab:experimental_datasets}
\begin{tabularx}{\textwidth}{|p{0.8cm}|p{0.7cm}|p{0.7cm}|p{1.3cm}|p{2.4cm}|p{1.6cm}|X|}
\hline

    \textbf{Dataset} & \textbf{Level} & \textbf{Avail?} & \textbf{Platform} & \textbf{Disorder} & \textbf{Classes} & \textbf{Statistics} \\
    \hline




    \cite{Shen2017-dy} & User & $\checkmark$ & $\mathbb{X}$ & Depression & Binary & Depressed Users: 1,402, Non-Depressed Users: $>$ 300 million, Depressed Tweets : 292,564, Non Depressed Tweets $>$ 10 billion \\ \hline
    \cite{10.1145/3308558.3313698} & User & $\checkmark$ & Reddit & Suicide Ideation & Multi-class & 500 total users; Attempt: 45; Behavior: 77; Ideation: 171; Indicator: 99; Supportive: 108. \\ \hline
    \cite{singh_twitter} & User & $\checkmark$ & $\mathbb{X}$ & ADHD, Bipolar, Anxiety, Depression, PTSD, OCD & Binary and Multi-class & 43269 tweets; 27003 users. \\ \hline
    \cite{yates-etal-2017-depression} & User & $\odot$ & Reddit & Depression & Binary & 9210 depressed users and 107274 control users. \\ \hline
    \cite{10.1007/978-3-030-45442-5_72} & User & $\odot$ & Reddit & Self-harm and depression & Binary and Multi-class & Submissions for Self-harm: 18,618, Submissions for non-self harm case: 254,642; Depression statistics: Not available, answers of BDI-II questionnaire \\ \hline
    \cite{10.1007/978-3-030-85251-1_22} & User & $\odot$ & Reddit & Pathological Gambling, Self-harm and Depression & Binary and Multi-class  & Gambling submissions: 54,674, Non-Gambling submissions: 1,073,883; Self-harm submissions: 69,722, Non self-harm submissions: 943,465; Depression statistics: Not available, answers of BDI-II questionnaire \\ \hline
    \cite{10.1007/978-3-031-13643-6_18} & User & $\odot$ & Reddit & Pathological Gambling, Self-harm and depression & Binary and Multi-class & Gambling submissions: 69,301, Non-Gambling submissions: 2,087,210; Depressed submissions: 35,332, Non depressed submissions: 687,228; Eating Disorders statistics: Not available, answers of EDE-Q questionnaire \\ \hline
    \cite{Pirayesh2021-zm} & User & $\odot$ & $\mathbb{X}$ & Depression & Binary & Depressed users: 372, Non-Depressed users: 445 \\ \hline
    \cite{10027637} & User & $\odot$ & $\mathbb{X}$ & Depression & Binary & Depressed users: 242, Non-Depressed users: 349 \\ \hline
    \cite{cohan-etal-2018-smhd} & User & $\odot$ & Reddit & Multiple (inc. Depression, Anxiety, Bipolar, OCD, Schizophrenia, ED) & Binary & Depression: 14,139 users and 1,272 posts; Anxiety: 8,783 users and 795K posts; Bipolar: 6,434 users and 575K posts; EDs: 598 users and 53K posts; OCD: 2,336 users and 203K posts; Schizophrenia: 1,331 users and 123K posts; Control: 335,952 users and 116M posts. \\ \hline
    \cite{mitchell-etal-2015-quantifying} & User & $\odot$ & $\mathbb{X}$ & Schizophrenia & Binary & 174 positively diagnosed users; 3200 posts. \\ \hline
    \cite{coppersmith-etal-2015-adhd} & User & $\odot$ & $\mathbb{X}$ & Multiple (inc. ADHD, Depression, Anxiety, OCD, ED, PTSD, Bipolar, Schizophrenia) & Binary & ADHD: 102 users and 384k posts; Anxiety: 216 users and 1591k posts; Bipolar 188 users and 720k posts; Depression: 393 users and 546k posts; EDs: 238 users and 724k posts; OCD: 100 users and 314k posts; PTSD: 403 users and 1251k posts; Schizophrenia: 172 users and 493k posts. \\ \hline
    \cite{10.1145/2858036.2858246} & User & $\odot$ & Tumblr & Recovery from Anorexia & Binary & 18,923 users and 55,334 posts. \\ \hline
    \cite{sekulic-etal-2018-just} & User & $\odot$ & Reddit & Bipolar Disorder & Binary & 3488 positively diagnosed users and 3931 control users. \\ \hline
    \cite{romero2024mentalriskes} & {User} & $\odot$ & Telegram & Depression,ED,Anxiety & Binary &\textbf{ED}: 335 subjects (192 Control, 143 Suffer) and 10,499 messages. \textbf{Depression:} 499 subjects (166 Control, 333 Suffer subclasses) and 17,370 messages. \textbf{Anxiety:} 500 subjects (57 Control, 443 Suffer) and 18,515 messages \\ \hline
    \cite{alhamed2024classifying} & {User} & $\checkmark$ & $\mathbb{X}$ & Depression & Binary & Before diagnosis: 250,785, after: 249,215, total 250 users. \\ \hline
    \cite{li2023mha} & {User} & $\odot$ & Weibo & Depression & Binary & Depressed: 2299, not depressed: 2307 \\ \hline

    \cite{guo2023leveraging} & {User} & $\odot$ & Weibo & Depression & Binary & Depressed:1570, Not depressed: 1570  \\ \hline
    \cite{abuhassan2025lexicon} & Post & $\checkmark$ & $\mathbb{X}$ & Eating Disorders & Multi-class & {Balanaced - None: 1250, Mild: 1250, Moderate: 1250, High: 1250; Imbalanced - None: 2750, Mild: 750, Moderate: 750, High: 750} \\ \hline
    \cite{Turcan2019-td} & Post & $\checkmark$ & Reddit & Depression & Binary & Depressed Posts: 1,857, Non-Depressed Posts: 1,698\\ \hline
    \cite{10.1145/3485447.3512128} & Post & $\checkmark$ & Reddit & Depression severity & Multi-class & Minimum depression level: 2,587, Mild depression level: 290, Moderate depression level: 394, Severe depression level: 282 \\ \hline
    \cite{ODEA2015183} & Post & $\odot$ & $\mathbb{X}$ & Suicide Ideation & Multi-class & 534 safe to ignore posts; 1029 Possibly concerning posts; 258 strongly concerning posts. \\ \hline
    \cite{rahman2024depressionemo} & {Post} & $\checkmark$ & Reddit & Depression Emotion & Multi-label & Sadness: 4,665, hopelessness: 4,193, worthlessness: 2,991, loneliness: 2,802, anger: 2,445, emptiness: 2,272, suicide intent: 1,491 \\ \hline

    \cite{beniwal2024hybrid} & {Post} & $\odot$ & Instagram & Depression & Binary & Depressive: 3431 posts and non-depressive: 6863 \\ \hline
    \cite{ghosh2023attention} & {Post} & $\odot$ & Facebook, $\mathbb{X}$, Youtube & Depression & Binary & Depressive: 2,525, Non-depressive Posts: 2,078 \\ \hline

\end{tabularx}
\end{table*}

\subsection{Evaluation measures}

Evaluating XAI models for MDD requires more than assessing predictive accuracy. While traditional metrics such as accuracy, F$_1$, and ROC-AUC remain essential, they do not capture whether predictions are clinically meaningful, timely, or aligned with expert reasoning. For health applications, evaluation must incorporate both task-specific performance measures and explainability-oriented properties. The metrics outlined below therefore serve two complementary purposes: (i) quantifying the reliability and faithfulness of model explanations, and (ii) assessing the quality and appropriateness of model outputs for MDD. {
In the context of explainable AI, an important distinction exists between faithfulness and plausibility. Faithfulness evaluates whether an explanation accurately reflects the model’s internal reasoning (sufficiency, comprehensiveness, and robustness), but it does not necessarily guarantee that the explanation is meaningful to human experts. In healthcare contexts such as MDD, explanations must also be plausible, meaning that they align with established clinical knowledge and appear reasonable to practitioners. Consequently, evaluation should consider both properties. In practice, plausibility is typically assessed through expert review, alignment with diagnostic frameworks (e.g., DSM-V), or human-centred evaluation protocols.
}

\subsubsection{Explainability-driven evaluation}

De Young et al.~\cite{deyoung-etal-2020-eraser} emphasise that explanations must be faithful, meaning that the features identified as important should directly support the model’s true internal reasoning. Faithfulness is commonly assessed using sufficiency and comprehensiveness. Let $\mathbf{x}$ be the model input. Let $M_I$ be a binary mask identifying features deemed \textit{important} by a given explanation method, and let $M_U = 1 - M_I$ denote the \textit{unimportant} features. The masked inputs are: 
\[
\mathbf{x}_I = \mathbf{x} \odot M_I, \qquad
\mathbf{x}_U = \mathbf{x} \odot M_U,
\]
and performance is evaluated by the F$_1$ score against the true label $y$.

\begin{definition}[Sufficiency ($\mathbf{Su}$)]
$\mathbf{Su}$ measures how well the model performs using only the important features:
\begin{equation}
\mathbf{Su} = 
F_1\!\left( \arg\max\left(f_{\theta}(\mathbf{x}_I)\right),\, y \right).
\end{equation}
High sufficiency indicates that the explanation captures most of the evidence required for the prediction.
\end{definition}

\begin{definition}[Comprehensiveness ($\mathbf{Co}$)]
$\mathbf{Co}$ measures model performance using only the unimportant features:
\begin{equation}
\mathbf{Co} =
F_1\!\left( \arg\max\left(f_{\theta}(\mathbf{x}_U)\right),\, y \right).
\end{equation}
Lower comprehensiveness indicates that removing important features meaningfully degrades the model's prediction.
\end{definition}

A second key property is robustness, which evaluates how stable explanations remain under input perturbations~\cite{holland2021robustness}. Let $\delta(\cdot)$ denote a perturbation operator (e.g., token masking, synonym replacement, random replacement, or Gaussian embedding noise), and define:
\[
\mathbf{x}_I^{\delta} = \delta(\mathbf{x}_I), \qquad 
\mathbf{x}_U^{\delta} = \delta(\mathbf{x}_U).
\]

\begin{definition}[Robustness]
Robustness evaluates how predictions change when important or unimportant features are perturbed:
\begin{align}
    \mathbf{Ro^+} &= F_1\!\left(\arg\max\left(f_{\theta}(\mathbf{x}_I^{\delta})\right),\, y\right),\\
    \mathbf{Ro^-} &= F_1\!\left(\arg\max\left(f_{\theta}(\mathbf{x}_U^{\delta})\right),\, y\right).
\end{align}
Perturbing important features ($\mathbf{Ro^+}$) should substantially disrupt the model’s performance (i.e., $\mathbf{Ro^+}$ should be low), whereas perturbing unimportant features ($\mathbf{Ro^-}$) should have minimal effect (i.e., $\mathbf{Ro^-}$ should remain high).
\end{definition}

\subsubsection{Task- and domain-specific evaluation}

Domain-specific measures are often used for specialised MDD scenarios, complementing general predictive metrics. These include early prediction (Definition \ref{def:erde}), predicting answers to MDD questionnaires (Definitions \ref{def:ahr}–\ref{def:adodl}), and multi-class classification of mental disorder severity (Definition \ref{def:ordinal}).

\begin{definition}[Early Risk Detection Error (ERDE)]\label{def:erde} ERDE~\cite{Losada2016-rd} penalises late correct predictions by incorporating the delay $k$ (number of posts observed before issuing an alert). A lower ERDE score indicates better early-detection performance.
\end{definition}

For a given instance:
\begin{numcases}{ERDE_{o}=} \label{eqn:erde}
  c_{fp} & \text{for False Positives (FP)}, \notag \\
  c_{fn} & \text{for False Negatives (FN)},  \\
  lc_{o}(k)\, c_{tp} & \text{for True Positives (TP)}, \notag \\
  0 & \text{for True Negatives (TN)}, \notag
\end{numcases}
where $lc_o(k) = 1 - \frac{1}{1 + \exp(k - o)}$.

\begin{definition}[Average Hit Rate (AHR)]\label{def:ahr}
AHR~\cite{PEREZ2022102380} measures the proportion of questionnaire items for which the model predicts the same response option as the user, averaged across users.
\end{definition}

\begin{definition}[Average Closeness Rate (ACR)]\label{def:acr}
Closeness Rate (CR) is defined as $\text{CR} = mad - ad$, where $mad$ is the maximum absolute difference and $ad$ is the observed absolute difference between predicted and actual questionnaire responses. ACR is the average CR across users~\cite{PEREZ2022102380}.
\end{definition}

\begin{definition}[Depression Category Hit Rate (DCHR)]\label{def:dchr}
DCHR computes the fraction of users for whom the predicted and actual questionnaire scores fall into the same severity band (e.g., BDI-II categories)~\cite{PEREZ2022102380}. Though introduced for depression, it generalises to other disorders with category-based questionnaires.
\end{definition}

\begin{definition}\textit{(Average Difference of Overall Depression Levels (ADODL))}\label{def:adodl}
The difference of overall depression levels (DODL) is defined as $\text{DODL} = \frac{63 - ad}{63}$, where $ad$ is the absolute difference between model-predicted and true BDI-II scores, and 63 is the maximum possible difference ($21$ items $\times$ $3$ points each). ADODL is the user-level average of DODL~\cite{PEREZ2022102380}.
\end{definition}

\begin{definition}[Ordinal Regression Metrics]\label{def:ordinal}
When severity labels have an inherent order (e.g., \textit{severe} $\rightarrow$ \textit{moderate} $\rightarrow$ \textit{mild} $\rightarrow$ \textit{minimum}), errors should reflect this structure. Following~\cite{10.1145/3437963.3441805}, ordinal false negatives and false positives are computed as:
\begin{equation}\label{eqn:fnfp}
    FN = \frac{\sum_{i=1}^{N_T} I(k_i^a > k_i^p)}{N_T}, \qquad
    FP = \frac{\sum_{i=1}^{N_T} I(k_i^a < k_i^p)}{N_T},
\end{equation}
where $k^a$ and $k^p$ denote the actual and predicted severity levels.
\end{definition}

\section{Research Horizons} \label{sec:researchhorizons}

\subsection{Open issues and challenges} \label{sec:issuesandchallenges}

The development of accurate, reliable, and explainable AI models for MDD in OSM environments faces several fundamental challenges. They affect data quality, model design, interpretability, and practical deployment, ultimately limiting the clinical and real-world impact.

{
\noindent\textbf{Dataset}. 
\textit{Issue:}
Benchmark datasets for MDD remain scarce, small-scale, and limited in scope. Most lack expert-annotated labels, fine-grained severity assessments, rich metadata (e.g., demographics and temporal markers), and user-level social network information \cite{zhang-etal-2023-sentiment, 10.1145/3485447.3512128}. The broader spectrum of mental disorders is also rarely represented, resulting in biased and incomplete modelling. OSM data are often unstructured, inconsistent, and noisy, with spurious correlations that can introduce bias during analysis \cite{olteanu2019social,giorgi2022correcting}.
\textit{Challenge:}
Improving dataset quality requires closer collaboration with domain experts to obtain clinically grounded annotations and implement bias mitigation strategies. Researchers also need approaches to overcome restricted API access in order to collect longitudinal data, user metadata, and social network structures that better reflect real-world mental health expressions. However, OSM data typically contain limited demographic information, which restricts transparency and the analysis of different user communities.
}

\noindent\textbf{User-level vs post-level MDD}.
\textit{Issue:}
While the ultimate goal is user-level MDD, much research continues to focus on post-level classification \cite{10027637,10.1145/3485447.3512128,10.1145/3437963.3441805}. Post-level analyses capture local emotional states but overlook long-term patterns, behavioural changes, and contextual factors that better reflect clinical MDD. \cite{10.1145/3543507.3583863}.
\textit{Challenge:}
Robust user-level detection demands integrative methods that incorporate users’ historical content, behavioural changes, and social interactions. Developing models that synthesise heterogeneous evidence over time, without overfitting to noisy or sparse signals, remains a significant methodological barrier.

{
\noindent\textbf{Binary classification vs severity estimation}. 
\textit{Issue:}
Most existing studies treat MDD as a binary outcome (i.e., presence or absence of symptoms), ignoring the continuum of depression severity \cite{Shen2017-dy, kuo2023dynamic}. This binary framing fails to support clinically meaningful tasks such as risk stratification or early intervention. The lack of severity analysis also constrains explainability because of limited insight into disorder-critical patterns such as specific symptoms or behavioural signals.
\textit{Challenge:}
Transitioning to fine-grained severity estimation requires standardised OSM-specific severity scales, high-quality longitudinal annotations, and models capable of identifying subtle linguistic and behavioural cues. Existing severity estimation research remains limited and largely disorder-specific \cite{10.1145/3485447.3512128, 9447025}. Generalisable, multi-disorder severity models are still absent. Furthermore, there is a lack of robust temporal and longitudinal assessment of individuals' mental states, which limits the possibility of timely interventions given the dynamic and episodic nature of mental disorders \cite{tsakalidis2022identifying}. Despite some efforts in relapse detection \cite{agarwal2025redepress} and mood-switch detection \cite{tsakalidis2022identifying}, studies examining symptom progression, relapse risk, and recovery dynamics remain scarce.
}


\noindent\textbf{Multimodality and social graph}.
\textit{Issue:}
Most MDD models rely solely on text \cite{10.1145/3485447.3512128,jiang-etal-2020-detection, 9349170,han-etal-2022-hierarchical}, overlooking other valuable modalities such as behavioural signals, posting patterns, imagery, metadata, and users’ social network structures. This results in narrow, context-poor assessments of mental state.
\textit{Challenge:}
Future systems must integrate multimodal inputs and social graph information using scalable and robust learning frameworks. This includes modelling interaction dynamics via GNNs and capturing user behaviours and contextual information alongside textual content \cite{10.1145/3543507.3583863,10.1145/3404835.3462938, 10027637, kuo2023dynamic}. Designing multimodal fusion architectures that remain interpretable is a major unresolved challenge.

\noindent\textbf{Volume of OSM data}. 
\textit{Issue:}
User-level MDD often requires analysing extensive historical timelines. Long text sequences, high posting frequency, and sparsely relevant content complicate model efficiency and risk information dilution.
\textit{Challenge:}
Methods such as summarisation, hierarchical encoding, and selective attention must be designed to condense information while preserving clinically salient signals \cite{8978072}. Striking the right compression-fidelity balance is particularly difficult in mental health contexts, where subtle cues carry significant meaning.


{
\noindent\textbf{Explainability and clinical transparency}. 
\textit{Issue:}
Explainability remains one of the most under-addressed aspects of MDD. Many SOTA models rely on complex deep learning architectures that behave as black boxes \cite{10027637,kuo2023dynamic}. In healthcare, opaque predictions undermine trust, limit adoption, and raise ethical concerns \cite{ibrahimov2025depressionx, Zogan2022,thamrin2025distinguishing}.
\textit{Challenge:}
A central challenge is achieving high predictive performance without sacrificing interpretability. Models must generate clinically meaningful explanations that identify relevant features, temporal drivers, and social-contextual factors while accounting for the complexity of multimodal and graph-structured data. Explanations produced by methods such as LIME, SHAP, attention mechanisms, and LLMs must also be clinically validated against established diagnostic frameworks such as the DSM-V.
}


\noindent\textbf{Temporal evolution of mental states}.
\textit{Issue:} Most current MDD research adopts a static perspective, treating mental health as a fixed snapshot rather than a dynamic process \cite{kuo2023dynamic}. This approach neglects fluctuations, relapse patterns, and transitions between mental states.
\textit{Challenge:}
Temporal modelling approaches capable of capturing short-term variability and long-term trajectories are needed. Dynamic deep neural networks, sequential GNNs, and time-aware transformers represent promising directions, but reliably modelling temporal mental health remains technically challenging.

\noindent\textbf{Simultaneous modelling of multiple co-occurring disorders}.
\textit{Issue:}
Existing studies typically detect a single disorder in isolation \cite{10.1145/3543507.3583863}, despite the well-established comorbidity of mental health conditions such as depression, anxiety, suicidality, and EDs \cite{jiang-etal-2020-detection}. Ignoring comorbidities oversimplifies real-world mental health and limits diagnostic value.
\textit{Challenge:}
Developing models capable of jointly modelling multiple, interrelated disorders requires unified representations of shared and disorder-specific cues, integration of temporal and multimodal evidence, and the incorporation of clinical knowledge describing disorder interdependencies.
This remains a major complex unresolved challenge in computational mental health.

\noindent\textbf{Integration of domain-specific mental health knowledge}.
\textit{Issue:}
Although some works incorporate lexicons, psychological features, or clinical questionnaires \cite{abuhassan2023classification,ibrahimov2025depressionx, 9906413,10.1145/3539618.3591905,PEREZ2022102380}, most models lack systematic and deep integration of domain expertise. This limits interpretability and clinical relevance.
\textit{Challenge:}
Interdisciplinary collaboration is essential to extract, formalise, and encode mental health knowledge into model architectures. Achieving this within deep learning frameworks, while maintaining transparency, requires new methods for knowledge grounding, explainable decision pathways, and model-clinician co-design.

{
\noindent\textbf{Ethical and legal aspects}.  
\textit{Issue:}  
Despite high predictive performance, existing studies often overlook ethical and legal considerations, including the consequences of false predictions, the absence of informed consent, and risks of re-identification or stigmatisation. Public social media profiles do not constitute formal consent for research participation, and false positives can result in unwanted interventions or worsening of symptoms \cite{neiders2025ethical,lane2022towards}.  
\textit{Challenge:}  
Ethical deployment of AI in healthcare is increasingly governed by regulatory frameworks. The EU AI Act classifies AI applications in healthcare as high-risk, requiring rigorous data governance, transparency, and meaningful human oversight \cite{AIActExplorer2024}. Similarly, WHO guidelines emphasise human-in-the-loop mechanisms to mitigate risks arising from automated tools \cite{WHO_LMM_2025}.
}


\subsection{Future research directions}	
\label{sec:directions}	

Explainable AI for MDD offers opportunities to enhance transparency, interpretability, and real-world impact, building on the challenges outlined above.

\noindent\textbf{Enhanced multi-modal data collection}.
Future work should prioritise the creation of richer, more representative multi-modal datasets that capture textual and multimedia posts, user interactions, behavioural signals, social networks, and contextual information \cite{Cao2019-fp,9308975}. Collaborations with mental health professionals are essential to ensure high-quality annotations, particularly for severity levels and temporal trajectories, that can better support fine-grained and clinically meaningful analyses \cite{9906413}.

\noindent\textbf{Transfer learning for multi-disorder detection}.
Mental disorders often co-occur, yet models are typically trained in isolation. Transfer learning offers a promising approach to leverage knowledge learned from one disorder to improve detection of others \cite{jiang-etal-2020-detection}. Advancing this direction requires systematic modelling of shared and disorder-specific patterns, enabling more adaptable and holistic mental health classifiers.

\noindent\textbf{Social graph representation learning}.
As social connections and interaction patterns strongly shape mental health expressions, future research should expand graph-based learning techniques to capture structural and temporal dynamics in online communities \cite{kuo2023dynamic, Fan2021-bm}. Novel approaches incorporating evolving relationships, contextual metadata, and multi-modal node attributes could provide deeper insights into disorder severity and co-evolving mental health states.

\noindent\textbf{Domain-aware fine-grained severity detection}.
A major gap remains in capturing fine-grained severity levels with explicit grounding in clinical knowledge. Future models should integrate symptom ontologies, validated clinical criteria, and expert-informed features \cite{9906413,10.1145/3539618.3591905,mitchell-etal-2015-quantifying,Yan2019-ih}. Crucially, such models must also provide transparent and clinically interpretable explanations to support practitioner trust and safe deployment.

\noindent\textbf{Real-time mental health analytics}.
Given the dynamic nature of mental health, research should shift towards real-time and longitudinal analytics capable of tracking moment-by-moment changes and detecting emerging risks. This involves developing adaptive models that respond to sudden shifts in online behaviour during major global events such as conflicts, recessions, or pandemics \cite{kuo2023dynamic, Losada2019-cm, 10.1007/978-3-031-13643-6_18}. Real-time systems may enable earlier, proactive, and more personalised interventions.

\noindent\textbf{Novel explainability methods}.
As GNNs and multi-modal architectures become more prominent, dedicated explainability techniques are needed to handle their complexity. Future methods should account for heterogeneous features, relational dependencies, and temporal histories \cite{Fan2021-bm,Ying2019-zg,Luo2020-tq}. Designing faithful, stable, and user-friendly explanations for such models is essential to improving transparency and enabling real-world adoption.


{
\noindent\textbf{Explanation with LLMs}.
Traditional XAI methods such as SHAP and LIME often fail to provide clinically relevant explanations for clinicians and patients \cite{kandala2025explainability,lee2025prompt}. LLMs can translate these outputs into human-readable, context-aware rationales \cite{zytek2024explingo}. When integrated with MDD pipelines via RAG, chain-of-thought reasoning, or specialised prompting \cite{Yang2023-yx,xu2024mental}, LLMs can generate clinically meaningful explanations for person-centered assessment \cite{lee2025prompt}. However, these explanations may be hallucinated or misleading, therefore require clinical validation, factual consistency checks, and human-in-the-loop oversight \cite{kim2503medical}. Future work should evaluate the faithfulness, stability, and clinical reliability of LLM-based explainability.
}

\noindent\textbf{Ethical considerations and bias mitigation}.
Ensuring fairness, privacy, and responsible use remains a core requirement. Research must address biases arising from imbalanced datasets, cultural skew, and algorithmic disparities, while also safeguarding sensitive user data \cite{MORLEY2020113172}. Ethical guidelines and transparent governance frameworks will be essential for real-world implementation

\noindent\textbf{Human-in-the-loop approaches}.
Rather than fully automated decision-making, future systems should incorporate human-in-the-loop mechanisms that involve mental health professionals during training, evaluation, and deployment \cite{zhou2025intermind,10.1145/3359249}. Expert feedback can enhance interpretability, reduce harmful errors, and ensure that explanations remain clinically meaningful.

\noindent\textbf{Incorporating cultural variations}.
Mental health expressions vary across cultural contexts, making cultural adaptability essential for global applicability. Future research should collaborate with domain experts from diverse regions to create culturally representative datasets and develop models capable of generalising across linguistic norms and cultural interpretations of distress \cite{10.1145/3359169,gopalkrishnan2018cultural}.

{\noindent\textbf{Extension to general clinical NLP contexts.} Although this survey focuses on OSM data, the reviewed AI techniques can also be applied to clinical transcripts, interview notes, lab results, and EHRs for mental disorder screening. Challenges such as model opacity, bias, and limited domain knowledge integration also arise in clinical contexts. In practice, these systems should serve as decision-support tools rather than standalone diagnostic systems. Explainable AI can assist clinicians by highlighting relevant linguistic cues and providing interpretable reasoning.
}


\section{Conclusion} 
\label{sec:conclusion}

There is an urgent need for effective tools for early detection and intervention in mental health. While existing AI models show strong predictive performance, their black-box nature raises ethical and practical concerns. This has driven interest in XAI to enhance transparency and interpretability in mental health AI. This survey reviewed traditional diagnostic approaches, AI methods, and XAI techniques, highlighting the importance of interpretable models for trust, accountability, and clinical relevance. Key future directions include real-time analytics, multimodal fusion, severity-aware modelling, graph-based explainability, and generative-model integration. The goal is AI systems that deliver accurate, actionable, and interpretable insights, supporting clinicians and individuals while improving societal well-being.





\bibliographystyle{IEEEtran}
\bibliography{references}

@inproceedings{jain2019attention,
  title={Attention is not Explanation},
  author={Jain, Sarthak and Wallace, Byron C},
  booktitle={NAACL:HLT},
  pages={3543--3556},
  year={2019}
}

@article{gao2025large,
  title={Large language model powered knowledge graph construction for mental health exploration},
  author={Gao, Shan and Yu, Kaixian and Yang, Yue and Yu, Sheng and Shi, Chenglong and Wang, Xueqin and Tang, Niansheng and Zhu, Hongtu},
  journal={Nature Communications},
  volume={16},
  number={1},
  pages={7526},
  year={2025}
}

@ARTICLE{thamrin2025distinguishing,
  author={Thamrin, Syauki Aulia and Chen, Arbee L.P.},
  journal={IEEE Trans. on Affective Computing}, 
  title={Distinguishing Depression and Bipolar Disorder from Social Media Data Utilizing Intensity of Emotions and Interpretable Deep Learning Models}, 
  year={2025},
  volume={},
  number={},
  pages={1-15},
  doi={10.1109/TAFFC.2025.3606887}}

@inproceedings{wang2025end,
  title={End-to-End Learnable Psychiatric Scale Guided Risky Post Screening for Depression Detection on Social Media},
  author={Wang, Bichen and Zi, Yuzhe and Sun, Yixin and Yang, Hao and Zhao, Yanyan and Qin, Bing},
  booktitle={EMNLP},
  pages={4054--4066},
  year={2025}
}

@inproceedings{liu2024depression,
  title={Depression detection via capsule networks with contrastive learning},
  author={Liu, Han and Li, Changya and Zhang, Xiaotong and Zhang, Feng and Wang, Wei and Ma, Fenglong and Chen, Hongyang and Yu, Hong and Zhang, Xianchao},
  booktitle={AAAI},
  volume={38},
  number={20},
  pages={22231--22239},
  year={2024}
}

@inproceedings{zhou2025intermind,
  title={InterMind: Doctor-Patient-Family Interactive Depression Assessment Empowered by Large Language Models},
  author={Zhou, Zhiyuan and Liu, Jilong and Wang, Sanwang and Hao, Shijie and Guo, Yanrong and Hong, Richang},
  booktitle={Proc. ACM Multimedia},
  pages={5480--5489},
  year={2025}
}

@article{xu2024mental,
  title={Mental-llm: Leveraging large language models for mental health prediction via online text data},
  author={Xu, Xuhai and Yao, Bingsheng and Dong, Yuanzhe and Gabriel, Saadia and Yu, Hong and Hendler, James and Ghassemi, Marzyeh and Dey, Anind K and Wang, Dakuo},
  journal={Proc. of the ACM on Interactive, Mobile, Wearable and Ubiquitous Technologies},
  volume={8},
  number={1},
  pages={1--32},
  year={2024}
}

@inproceedings{ibrahimov2025depressionx,
  title={DepressionX: Knowledge Infused Residual Attention for Explainable Depression Severity Assessment},
  author={Ibrahimov, Yusif and Anwar, Tarique and Yuan, Tommy},
  booktitle={Proc. of the Workshop on Health Intelligence (W3PHIAI), in Conjunction with AAAI},
  year={2025}
}

@ARTICLE{abuhassan2025lexicon,
author={Abuhassan, Mohammad and Anwar, Tarique and Liu, Chengfei and Jarman, Hannah K. and McLean, Sian and Paxton, Susan and Rodgers, Rachel F. and Fuller-Tyszkiewicz, Matthew},
journal={ IEEE Trans. on Affective Computing },
title={{ Lexicon-Based Graph Attention for Severity Estimation of Eating Disorders on Social Media }},
year={2025},
ISSN={1949-3045},
pages={1-14},
doi={10.1109/TAFFC.2025.3628354}
}

@MISC{who_mental_disorders,
year = {2024},
	title = {Mental Disorders},
	editor = {World Health Organization},
organization = {WHO Fact Sheets},
	url = {https://www.who.int/news-room/fact-sheets/detail/mental-disorders},
 }

@article{ZHANG2023231,
title = {Emotion fusion for mental illness detection from Social media: A survey},
journal = {Inf. Fus.},
volume = {92},
pages = {231-246},
year = {2023},
author = {Tianlin Zhang and Kailai Yang and Shaoxiong Ji and Sophia Ananiadou},
}

@article{10.1145/3572406,
author = {Ahmed, Usman and Lin, Jerry Chun-Wei and Srivastava, Gautam},
title = {Graph Attention Network for Text Classification and Detection of Mental Disorder},
year = {2023},
volume = {17},
number = {3},
journal = {ACM Trans. Web},
pages={1--31},

}

@article{PEREZ2022102380,
title = {Automatic depression score estimation with word embedding models},
journal = {AI in Med.},
volume = {132},
pages = {102380},
year = {2022},
author = {Anxo Pérez and Javier Parapar and Álvaro Barreiro},
}

@article{Thompson2004,
  author = {Thompson, Anna and Hunt, Caroline and Issakidis, Cathy},
  title = {Why wait? Reasons for delay and prompts to seek help for mental health problems in an Australian clinical sample},
  journal = {Social Psychiatry and Psychiatric Epidem.},
  volume = {39},
  number = {10},
  pages = {810--817},
  year = {2004}
}

@inproceedings{10.1145/3543507.3583863,
author = {Abuhassan, Mohammad and Anwar, Tarique and Liu, Chengfei and Jarman, Hannah K and Fuller-Tyszkiewicz, Matthew},
title = {EDNet: Attention-Based Multimodal Representation for Classification of Twitter Users Related to Eating Disorders},
year = {2023},
booktitle = {ACM Web Conf.},
pages = {4065–4074}
}

@ARTICLE{9447025,
  author={Ghosh, Shreya and Anwar, Tarique},
  journal={IEEE Trans. on Comput. Soc. Syst.}, 
  title={Depression Intensity Estimation via Social Media: A Deep Learning Approach}, 
  year={2021},
  volume={8},
  number={6},
  pages={1465-1474}}

@article{Zhang2022-gm,
  author = {Zhang, Tianlin and Schoene, Annika M. and Ji, Shaoxiong and Ananiadou, Sophia},
  title = {Natural language Processing applied to mental illness detection: a narrative review},
  journal = {npj Dig. Med.},
  volume = {5},
  number = {1},
  pages = {46},
  year = {2022}
}

@inproceedings{10.1145/3485447.3512128,
author = {Naseem, Usman and Dunn, Adam G. and Kim, Jinman and Khushi, Matloob},
title = {Early Identification of Depression Severity Levels on Reddit Using Ordinal Classification},
year = {2022},
booktitle = {ACM Web Conf.},
pages = {2563–2572}
}

@article{DALFONSO2020112,
title = {AI in mental health},
journal = {Current Opinion in Psychology},
volume = {36},
pages = {112-117},
year = {2020},
author = {Simon D’Alfonso},
}

@article{10.1145/3471902,
author = {Chen, Min and Shen, Ke and Wang, Rui and Miao, Yiming and Jiang, Yingying and Hwang, Kai and Hao, Yixue and Tao, Guangming and Hu, Long and Liu, Zhongchun},
title = {Negative Information Measurement at AI Edge: A New Perspective for Mental Health Monitoring},
year = {2022},
volume = {22},
number = {3},
journal = {ACM Trans. Internet Technol.},
pages={1--16},
}

@online{datareportal_social_media_users,
  organization =  {DataReportal},
  title =         {Social Media Users},
  url =           {https://datareportal.com/social-media-users}
}

@inproceedings{10.1145/3539618.3591905,
author = {P\'{e}rez, Anxo and Parapar, Javier and Barreiro, \'{A}lvaro and Lopez-Larrosa, Silvia},
title = {BDI-Sen: A Sentence Dataset for Clinical Symptoms of Depression},
year = {2023},
booktitle = {ACM SIGIR},
pages = {2996–3006}
}

@inproceedings{10.1145/3404835.3462938,
author = {Zogan, Hamad and Razzak, Imran and Jameel, Shoaib and Xu, Guandong},
title = {DepressionNet: Learning Multi-Modalities with User Post Summarization for Depression Detection on Social Media},
year = {2021},
booktitle = {ACM SIGIR},
pages = {133–142}
}

@inproceedings{bucur2021psychologically,
  title={A Psychologically Informed Part-of-Speech Analysis of Depression in Social Media},
  author={Bucur, Ana-Maria and Podin{\u{a}}, Ioana R and Dinu, Liviu P},
  booktitle={RANLP},
  pages={199--207},
  year={2021}
}

@inproceedings{coppersmith-etal-2016-exploratory,
    title = "Exploratory Analysis  of Social Media Prior to a Suicide Attempt",
    author = "Coppersmith, Glen  and
      Ngo, Kim  and
      Leary, Ryan  and
      Wood, Anthony",
    booktitle = "CLPsych",
    year = "2016",
    pages = "106--117",
}

@article{Zogan2022-wd,
  author = {Zogan, Hamad and Razzak, Imran and Wang, Xianzhi and Jameel, Shoaib and Xu, Guandong},
  title = {Explainable depression detection with multi-aspect features using a hybrid deep learning model on Social media},
  journal = {World Wide Web},
  volume = {25},
  number = {1},
  pages = {281--304},
  year = {2022}
}

@ARTICLE {9769937,
author = {Y. Jia and J. McDermid and T. Lawton and I. Habli},
journal = {IEEE T. on Emerg. Topics in Comput.},
title = {The Role of Explainability in Assuring Safety of Machine Learning in Healthcare},
year = {2022},
volume = {10},
number = {04},
pages = {1746-1760}
}

@book{DSM5,
  author = {American Psychiatric Association},
  title = {Diagnostic and statistical manual of mental disorders: {DSM-5™}, 5th ed.},
  year = {2013},
  publisher = {American Psychiatric Publishing, Inc.},
  pages = {xliv, 947}
}

@article{Beck1961-qd,
  author = {Beck, A. T. and Ward, C. H. and Mendelson, M. and Mock, J. and Erbaugh, J.},
  title = {An inventory for measuring depression.},
  journal = {Archives of General Psychiatry},
  volume = {4},
  pages = {561-571},
  year = {1961}
}

@article{doi:10.1177/014662167700100306,
author = {Lenore Sawyer Radloff},
title ={The CES-D Scale: A Self-Report Depression Scale for Research in the General Population},
journal = {App. Psycho. Meas.},
volume = {1},
number = {3},
pages = {385-401},
year = {1977}}

@article{beck_ii,
author = {Aaron T. Beck, Robert A. Steer, Roberta Ball and William F. Ranieri},
title = {Comparison of Beck Depression Inventories-IA and-II in Psychiatric Outpatients},
journal = {Journal of Personality Assessment},
volume = {67},
number = {3},
pages = {588-597},
year = {1996}}

@inproceedings{10.5555/3294771.3294869,
author = {Hamilton, William L. and Ying, Rex and Leskovec, Jure},
title = {Inductive Representation Learning on Large Graphs},
year = {2017},
booktitle = {NeurIPS},
pages = {1025–1035}
}

@article{Kroenke2001-ul,
  author = {Kroenke, Kurt and Spitzer, Robert L. and Williams, Janet B. W.},
  title = {The PHQ-9: Validity of a brief depression severity measure.},
  journal = {Journal of General Internal Med.},
  volume = {16},
  number = {9},
  pages = {606-613},
  year = {2001}
}

@article{10.1001/archinte.166.10.1092,
    author = {Spitzer, Robert L. and Kroenke, Kurt and Williams, Janet B. W. and Löwe, Bernd},
    title = "{A Brief Measure for Assessing Generalized Anxiety Disorder: The GAD-7}",
    journal = {Archives of Internal Med.},
    volume = {166},
    number = {10},
    pages = {1092-1097},
    year = {2006}
}

@incollection{Beck_undated-xw,
  author = {Steer, Robert A. and Beck, Aaron T.},
  title = {Beck Anxiety Inventory},
  booktitle = {Evaluating stress: A book of resources},
  year = {1997},
  publisher = {Scarecrow Education},
  pages = {23-40}
}

@article{Cooper1987-ov,
author = {Cooper, Zafra and Fairburn, Christopher},
title = {The eating disorder examination: A semi-structured interview for the assessment of the specific psychopathology of eating disorders},
journal = {Int. J. of Eat. Dis.},
volume = {6},
number = {1},
pages = {1-8},
year = {1987}
}

@article{garner_garfinkel_1979, title={The Eating Attitudes Test: an index of the symptoms of anorexia nervosa}, volume={9}, number={2}, journal={Psychological Med.}, author={Garner, David M. and Garfinkel, Paul E.}, year={1979}, pages={273–279}}

@article{10.1145/3589784,
author = {Dhelim, Sahraoui and Chen, Liming and Das, Sajal K. and Ning, Huansheng and Nugent, Chris and Leavey, Gerard and Pesch, Dirk and Bantry-White, Eleanor and Burns, Devin},
title = {Detecting Mental Distresses Using Social Behavior Analysis in the Context of COVID-19: A Survey},
year = {2023},
volume = {55},
number = {14s},
journal = {ACM Comput. Surv.},
articleno = {318},
}

@article{KADKHODA2022101042,
title = {Bipolar disorder detection over Social media},
journal = {Informatics in Med. Unlocked},
volume = {32},
pages = {101042},
year = {2022},
author = {Elham Kadkhoda and Mahsa Khorasani and Fatemeh Pourgholamali and Mohsen Kahani and Amir Rezaei Ardani}}

@ARTICLE{9906413,
  author={Anwar, Tarique and Fuller-Tyszkiewicz, Matthew and Jarman, Hannah K and Abuhassan, Mohammad and Shatte, Adrian and Team, WIRED and Sukunesan, Suku},
  journal={IEEE J. of Biomed. and Health Inform.}, 
  title={EDBase: Generating a Lexicon Base for Eating Disorders Via Social Media}, 
  year={2022},
  volume={26},
  number={12},
  pages={6116-6125}}

@article{kuo2023dynamic,
  title={Dynamic Graph Representation Learning for Depression Screening with Transformer},
  author={Kuo, Ai-Te and Chen, Haiquan and Kuo, Yu-Hsuan and Ku, Wei-Shinn},
  journal={arXiv:2305.06447},
  year={2023}
}

@article{Zogan2022,
  author = {Zogan, Hamad and Razzak, Imran and Wang, Xianzhi and Jameel, Shoaib and Xu, Guandong},
  title = {Explainable Depression Detection with Multi-Aspect Features Using a Hybrid Deep Learning Model on Social Media},
  journal = {World Wide Web},
  volume = {25},
  number = {1},
  pages = {281-304},
  year = {2022}
}

@INPROCEEDINGS{10027637,
  author={Mihov, Ivan and Chen, Haiquan and Qin, Xiao and Ku, Wei-Shinn and Yan, Da and Liu, Yuhong},
  booktitle={IEEE ICDM}, 
  title={MentalNet: Heterogeneous Graph Representation for Early Depression Detection}, 
  year={2022},
  volume={},
  number={},
  pages={1113-1118}}

@INPROCEEDINGS{8978072,
  author={Naseem, Usman and Musial, Katarzyna},
  booktitle={ICDAR}, 
  title={DICE: Deep Intelligent Contextual Embedding for Twitter Sentiment Analysis}, 
  year={2019},
  volume={},
  number={},
  pages={953-958}}

@article{Richter2020-jn,
  author = {Richter, Thalia and Fishbain, Barak and Markus, Andrey and Richter-Levin, Gal and Okon-Singer, Hadas},
  title = {Using Machine Learning-Based Analysis  for Behavioral Differentiation Between Anxiety and Depression},
  journal = {Scientific Reports},
  volume = {10},
  number = {1},
  pages = {16381},
  year = {2020}
}

@inproceedings{mitchell-etal-2015-quantifying,
    title = "Quantifying the Language of Schizophrenia in Social Media",
    author = "Mitchell, Margaret  and
      Hollingshead, Kristy  and
      Coppersmith, Glen",
    booktitle = "CLPsych",
    year = "2015",
    pages = "11--20",
}

@ARTICLE{9308975,
  author={Cao, Lei and Zhang, Huijun and Feng, Ling},
  journal={IEEE Trans. on Multimedia}, 
  title={Building and Using Personal Knowledge Graph to Improve Suicidal Ideation Detection on Social Media}, 
  year={2022},
  volume={24},
  number={},
  pages={87-102}}

@article{doi:10.1177/0276236615598957,
author = {Rachelle Pavelko and Jessica Gall Myrick},
title ={Tweeting and Trivializing: How the Trivialization of Obsessive–Compulsive Disorder via Social Media Impacts User Perceptions, Emotions, and Behaviors},
journal = {Imagination, Cognition and Personality},
volume = {36},
number = {1},
pages = {41-63},
year = {2016}}

@inproceedings{han-etal-2022-hierarchical,
    title = "Hierarchical Attention Network for Explainable Depression Detection on {T}witter Aided by Metaphor Concept Mappings",
    author = "Han, Sooji  and
      Mao, Rui  and
      Cambria, Erik",
    booktitle = "COLING",
    year = "2022",
    pages = "94--104",
}

@ARTICLE{9349170,
  author={Schoene, Annika Marie and Turner, Alexander P. and De Mel, Geeth and Dethlefs, Nina},
  journal={IEEE T. on Affect. Comput.}, 
  title={Hierarchical Multiscale Recurrent Neural Networks for Detecting Suicide Notes}, 
  year={2023},
  volume={14},
  number={1},
  pages={153-164}}

@InProceedings{10.1007/978-3-030-45442-5_72,
author="Losada, David E.
and Crestani, Fabio
and Parapar, Javier",
title="eRisk 2020: Self-harm and Depression Challenges",
booktitle="Adv. in Inf. Retr.",
year="2020",
pages="557--563",
}

@inproceedings{10.1007/978-3-030-85251-1_22,
  title={Overview of erisk 2023: Early risk prediction on the internet},
  author={Parapar, Javier and Mart{\'\i}n-Rodilla, Patricia and Losada, David E and Crestani, Fabio},
  booktitle={CLEF},
  pages={294--315},
  year={2023}
}

@inproceedings{10.1007/978-3-031-13643-6_18,
  title={Overview of erisk 2023: Early risk prediction on the internet},
  author={Parapar, Javier and Mart{\'\i}n-Rodilla, Patricia and Losada, David E and Crestani, Fabio},
  booktitle={CLEF},
  pages={294--315},
  year={2023},
  organization={Springer}
}

@inproceedings{10.1145/3308558.3313698,
author = {Gaur, Manas and Alambo, Amanuel and Sain, Joy Prakash and Kursuncu, Ugur and Thirunarayan, Krishnaprasad and Kavuluru, Ramakanth and Sheth, Amit and Welton, Randy and Pathak, Jyotishman},
title = {Knowledge-Aware Assessment of Severity of Suicide Risk for Early Intervention},
year = {2019},
booktitle = {The Web Conf.},
pages = {514–525},
}

@article{singh_twitter, 
title={Twitter-STMHD: An Extensive User-Level Database of Multiple Mental Health Disorders}, 
volume={16}, 
number={1}, 
journal={ICWSM}, 
author={, Suhavi and Singh, Asmit Kumar and Arora, Udit and Shrivastava, Somyadeep and Singh, Aryaveer and Shah, Rajiv Ratn and Kumaraguru, Ponnurangam}, 
year={2022}, 
pages={1182-1191} }

@inproceedings{yates-etal-2017-depression,
    title = "Depression and Self-Harm Risk Assessment in Online Forums",
    author = "Yates, Andrew  and
      Cohan, Arman  and
      Goharian, Nazli",
    booktitle = "EMNLP",
    year = "2017",
    pages = "2968--2978",
}

@inproceedings{jiang-etal-2020-detection,
    title = "Detection of Mental Health from {R}eddit via Deep Contextualized Representations",
    author = "Jiang, Zhengping  and
      Levitan, Sarah Ita  and
      Zomick, Jonathan  and
      Hirschberg, Julia",
    booktitle = "LOUHI",
    year = "2020",
    publisher = "ACL",
    pages = "147--156",
}

@inproceedings{cohan-etal-2018-smhd,
    title = "{SMHD}: a Large-Scale Resource for Exploring Online Language Usage for Multiple Mental Health Conditions",
    author = "Cohan, Arman  and
      Desmet, Bart  and
      Yates, Andrew  and
      Soldaini, Luca  and
      MacAvaney, Sean  and
      Goharian, Nazli",
    booktitle = "COLING",
    year = "2018",
    pages = "1485--1497"
}

@article{ODEA2015183,
title = {Detecting suicidality on Twitter},
journal = {Internet Interventions},
volume = {2},
number = {2},
pages = {183-188},
year = {2015},
author = {Bridianne O'Dea and Stephen Wan and Philip J. Batterham and Alison L. Calear and Cecile Paris and Helen Christensen}
}

@inproceedings{10.1145/3437963.3441805,
author = {Sawhney, Ramit and Joshi, Harshit and Gandhi, Saumya and Shah, Rajiv Ratn},
title = {Towards Ordinal Suicide Ideation Detection on Social Media},
year = {2021},
booktitle = {ACM WSDM},
pages = {22–30}
}

@inproceedings{zhang-etal-2023-sentiment,
    title = "Sentiment-guided Transformer with Severity-aware Contrastive Learning for Depression Detection on Social Media",
    author = "Zhang, Tianlin  and
      Yang, Kailai  and
      Ananiadou, Sophia",
    booktitle = "BioNLP",
    year = "2023",
    pages = "114--126",
}

@inproceedings{sekulic-etal-2018-just,
    title = "Not Just Depressed: Bipolar Disorder Prediction on {R}eddit",
    author = "Sekulic, Ivan  and
      Gjurkovi{\'c}, Matej  and
      {\v{S}}najder, Jan",
    booktitle = "WASSA",
    year = "2018",
    publisher = "ACL",
    pages = "72--78",
}

@inproceedings{10.1145/2858036.2858246,
author = {Chancellor, Stevie and Mitra, Tanushree and De Choudhury, Munmun},
title = {Recovery Amid Pro-Anorexia: Analysis of Recovery in Social Media},
year = {2016},
booktitle = {ACM CHI},
pages = {2111–2123},
}

@inproceedings{coppersmith-etal-2015-adhd,
    title = "From {ADHD} to {SAD}: Analyzing the Language of Mental Health on {T}witter through Self-Reported Diagnoses",
    author = "Coppersmith, Glen  and
      Dredze, Mark  and
      Harman, Craig  and
      Hollingshead, Kristy",
    booktitle = "CLPsych",
    year = "2015",
    pages = "1--10",
}

@inproceedings{10.1145/3136755.3136766,
author = {Bhatia, Shalini and Hayat, Munawar and Goecke, Roland},
title = {A Multimodal System to Characterise Melancholia: Cascaded Bag of Words Approach},
year = {2017},
booktitle = {ACM ICMI},
pages = {274–280},
}

@INPROCEEDINGS{7752261,
  author={Chang, Chun-Hao and Saravia, Elvis and Chen, Yi-Shin},
  booktitle={IEEE/ACM ASONAM}, 
  title={Subconscious Crowdsourcing: A feasible data collection mechanism for mental disorder detection on Social media}, 
  year={2016},
  pages={374-379}}

@INPROCEEDINGS{7752434,
  author={Saravia, Elvis and Chang, Chun-Hao and De Lorenzo, Renaud Jollet and Chen, Yi-Shin},
  booktitle={IEEE/ACM ASONAM}, 
  title={MIDAS: Mental illness detection and Analysis  via Social media}, 
  year={2016},
  volume={},
  number={},
  pages={1418-1421}}

@article{Yan2019-ih,
  author = {Yan, Hao and Fitzsimmons-Craft, Ellen E and Goodman, Micah and Krauss, Melissa and Das, Sanmay and Cavazos-Rehg, Patricia},
  title = {Automatic Detection of Eating Disorder-Related Social Media Posts That Could Benefit From a Mental Health Intervention},
  journal = {Int. J. of Eat. Dis.},
  volume = {52},
  number = {10},
  pages = {1150-1156},
  year = {2019}
}

@inproceedings{Shickel2016-su,
    title = "Automatic Triage of Mental Health Forum Posts",
    author = "Shickel, Benjamin  and
      Rashidi, Parisa",
    booktitle = "CLPsych",
    year = "2016",
    pages = "188--192",
}

@inproceedings{10.1145/3442536.3442553,
author = {Skaik, Ruba and Inkpen, Diana},
title = {Using Twitter Social Media for Depression Detection in the Canadian Population},
year = {2021},
booktitle = {AICCC},
pages = {109–114},
}

@inproceedings{10.1145/3543507.3583350,
author = {Yuan, Yunhao and Saha, Koustuv and Keller, Barbara and Isomets\"{a}, Erkki Tapio and Aledavood, Talayeh},
title = {Mental Health Coping Stories on Social Media: A Causal-Inference Study of Papageno Effect},
year = {2023},
booktitle = {ACM Web Conf.},
pages = {2677–2685},
}

@INPROCEEDINGS{9918782,
  author={Lim, Yan Qian and Lee, Ming Jie and Loo, Yim Ling},
  booktitle={AiDAS}, 
  title={Towards A Machine Learning Framework for Suicide Ideation Detection in Twitter}, 
  year={2022},
  volume={},
  number={},
  pages={153-157}}

@ARTICLE{9416889,
  author={Aragón, Mario Ezra and López-Monroy, Adrian Pastor and González-Gurrola, Luis Carlos and Montes-y-Gómez, Manuel},
  journal={IEEE Trans. on Affect. Comput.}, 
  title={Detecting Mental Disorders in Social Media Through Emotional Patterns - The Case of Anorexia and Depression}, 
  year={2023},
  volume={14},
  number={1},
  pages={211-222}}

@article{Ma2023-ts,
title = {An integrated latent Dirichlet allocation and Word2vec method for generating the topic evolution of mental models from global to local},
journal = {Expert Syst. with Applic.},
volume = {212},
pages = {118695},
year = {2023},
author = {Jian Ma and Lei Wang and Yuan-Rong Zhang and Wei Yuan and Wei Guo},
}

@Article{Hagg2022-tb,
author="Hagg, Lauryn J
and Merkouris, Stephanie S
and O'Dea, Gypsy A
and Francis, Lauren M
and Greenwood, Christopher J
and Fuller-Tyszkiewicz, Matthew
and Westrupp, Elizabeth M
and Macdonald, Jacqui A
and Youssef, George J",
title="Examining Analytic Practices in Latent Dirichlet Allocation Within Psychological Science: Scoping Review",
journal="J. Med. Internet Res.",
year="2022",
volume="24",
number="11",
pages="e33166",
}

@article{Lee2020-ps,
author = {Sherman A. Lee},
title = {Coronavirus Anxiety Scale: A brief mental health screener for COVID-19 related anxiety},
journal = {Death Studies},
volume = {44},
number = {7},
pages = {393-401},
year = {2020}}

@inproceedings{Pirayesh2021-zm,
author = {Pirayesh, Jahandad and Chen, Haiquan and Qin, Xiao and Ku, Wei-Shinn and Yan, Da},
title = {MentalSpot: Effective Early Screening for Depression Based on Social Contagion},
year = {2021},
booktitle = {ACM CIKM},
pages = {1437–1446}
}

@inproceedings{Turcan2019-td,
    title = "{D}readdit: A {R}eddit Dataset for Stress Analysis  in Social Media",
    author = "Turcan, Elsbeth  and
      McKeown, Kathy",
    booktitle = "LOUHI",
    year = "2019",
    publisher = "ACL",
    pages = "97--107",
}

@inproceedings{Shen2017-dy,
  author    = {Guangyao Shen and Jia Jia and Liqiang Nie and Fuli Feng and Cunjun Zhang and Tianrui Hu and Tat-Seng Chua and Wenwu Zhu},
  title     = {Depression Detection via Harvesting Social Media: A Multimodal Dictionary Learning Solution},
  booktitle = {IJCAI},
  pages     = {3838--3844},
  year      = {2017},
}

@inproceedings{Losada2016-rd,
  title={A test collection for research on depression and language use},
  author={Losada, David E and Crestani, Fabio},
  booktitle={CLEF},
  pages={28--39},
  year={2016}
}

@inproceedings{Yang2023-yx,
  title={Towards interpretable mental health analysis with large language models},
  author={Yang, Kailai and Ji, Shaoxiong and Zhang, Tianlin and Xie, Qianqian and Kuang, Ziyan and Ananiadou, Sophia},
  booktitle={EMNLP},
  pages={6056--6077},
  year={2023}
}

@inproceedings{Coppersmith2014-kw,
    title = "Quantifying Mental Health Signals in {T}witter",
    author = "Coppersmith, Glen  and
      Dredze, Mark  and
      Harman, Craig",
    booktitle = "CLPsych",
    year = "2014",
    pages = "51--60",
}

@article{Mikolov2013-kq,
  title={Efficient estimation of word representations in vector space},
  author={Mikolov, Tomas and Chen, Kai and Corrado, Greg and Dean, Jeffrey},
  journal={arXiv:1301.3781},
  year={2013}
}

@inproceedings{Arya2019-yg,
  title={One Explanation Does Not Fit All: A Toolkit And Taxonomy Of AI Explainability Techniques},
  author={Arya, Vijay and Bellamy, Rachel K and Chen, Pin-Yu and Dhurandhar, Amit and Hind, Michael and Hoffman, Samuel C and Houde, Stephanie and Liao, Q Vera and Luss, Ronny and Mojsilovic, Aleksandra and others},
  booktitle={INFORMS Ann. Meeting},
  year={2021}
}

@article{Straw2020-lp,
    author = {Straw, Isabel AND Callison-Burch, Chris},
    journal = {PLOS ONE},
    title = {AI in mental health and the biases of language based models},
    year = {2020},
    volume = {15},
    pages = {1-19},
    number = {12},

}

@article{Liu2022-qx,
  author = {Liu, Yu and Xu, Chen and Kuai, Xi and Deng, Hao and Wang, Kaifeng and Luo, Qinyao},
  title = {Analysis  of the Causes of Inferiority Feelings Based on Social Media Data with Word2Vec},
  journal = {Scientific Reports},
  volume = {12},
  number = {1},
  pages = {5218},
  year = {2022},
}

@inproceedings{Pennington2014-fc,
  title={Glove: Global vectors for word representation},
  author={Pennington, Jeffrey and Socher, Richard and Manning, Christopher D},
  booktitle={EMNLP},
  pages={1532--1543},
  year={2014}
}

@article{Ahmed2022-oy,
title = {EANDC: An explainable attention network based deep adaptive clustering model for mental health treatment},
journal = {Future Generation Computer Syst.},
volume = {130},
pages = {106-113},
year = {2022},
author = {Usman Ahmed and Gautam Srivastava and Unil Yun and Jerry Chun-Wei Lin}
}

@article{Dheeraj2021-at,
title = {Negative emotions detection on online mental-health related patients texts using the deep learning with MHA-BCNN model},
journal = {Expert Syst. with Applications},
volume = {182},
pages = {115265},
year = {2021},
author = {Kodati Dheeraj and Tene Ramakrishnudu},
}

@inproceedings{Devlin2018-zf,
    title = "{BERT}: Pre-training of Deep Bidirectional Transformers for Language Understanding",
    author = "Devlin, Jacob  and
      Chang, Ming-Wei  and
      Lee, Kenton  and
      Toutanova, Kristina",
    booktitle = "NAACL",
    year = "2019",
    pages = "4171--4186",
}

@inproceedings{bayram-benhiba-2022-emotionally,
    title = "Emotionally-Informed Models for Detecting Moments of Change and Suicide Risk Levels in Longitudinal Social Media Data",
    author = "Bayram, Ulya  and
      Benhiba, Lamia",
    booktitle = "CLPsych",
    year = "2022",
    pages = "219--225",
}

@inproceedings{Yao2019-vx,
  title={Graph convolutional networks for text classification},
  author={Yao, Liang and Mao, Chengsheng and Luo, Yuan},
  booktitle={AAAI},
  volume={33},
  number={01},
  pages={7370--7377},
  year={2019}
}

@article{abuhassan2023classification,
  title={Classification of Twitter users with eating disorder engagement: Learning from the biographies},
  author={Abuhassan, Mohammad and Anwar, Tarique and Fuller-Tyszkiewicz, Matthew and Jarman, Hannah K and Shatte, Adrian and Liu, Chengfei and Sukunesan, Suku},
  journal={Computers in Human Behav.},
  volume={140},
  pages={107519},
  year={2023}
}

@article{anwar2023tracking,
  title={Tracking the Evolution of Clusters in Social Media Streams},
  author={Anwar, Tarique and Nepal, Surya and Paris, Cecile and Yang, Jian and Wu, Jia and Sheng, Quan Z},
  journal={IEEE T. on Big Data},
  volume={9},
  number={2},
  pages={701--715},
  year={2023}
}

@inproceedings{Liu2020-ah,
  title={Tensor graph convolutional networks for text classification},
  author={Liu, Xien and You, Xinxin and Zhang, Xiao and Wu, Ji and Lv, Ping},
  booktitle={AAAI},
  volume={34},
  number={05},
  pages={8409--8416},
  year={2020}
}

@inproceedings{
Kipf2016-af,
title={Semi-Supervised Classification with Graph Convolutional Networks},
author={Thomas N. Kipf and Max Welling},
booktitle={ICLR},
year={2017}
}

@article{Rosenquist2011-jx,
  author = {Rosenquist, J. N. and Fowler, J. H. and Christakis, N. A.},
  title = {Social Network Determinants of Depression},
  journal = {Molecular Psychiatry},
  volume = {16},
  number = {3},
  pages = {273--281},
  year = {2011}
}

@ARTICLE{Zogan2023-by,
  author={Zogan, Hamad and Razzak, Imran and Jameel, Shoaib and Xu, Guandong},
  journal={IEEE J. of Biomed. and Health Inform.}, 
  title={Hierarchical Convolutional Attention Network for Depression Detection on Social Media and Its Impact During Pandemic}, 
  year={2023},
  volume={},
  number={},
  pages={1-9}}

@article{Hochreiter1997-jc,
    author = {Hochreiter, Sepp and Schmidhuber, Jürgen},
    title = "{Long Short-Term Memory}",
    journal = {Neural Computation},
    volume = {9},
    number = {8},
    pages = {1735-1780},
    year = {1997}
}

@inproceedings{Gaur2019-tr,
author = {Gaur, Manas and Alambo, Amanuel and Sain, Joy Prakash and Kursuncu, Ugur and Thirunarayan, Krishnaprasad and Kavuluru, Ramakanth and Sheth, Amit and Welton, Randy and Pathak, Jyotishman},
title = {Knowledge-Aware Assessment of Severity of Suicide Risk for Early Intervention},
year = {2019},
booktitle = {The Web Conf.},
pages = {514–525},
numpages = {12}
}

@INPROCEEDINGS{hu2018squeeze,
  author={Hu, Jie and Shen, Li and Sun, Gang},
  booktitle={IEEE/CVF CVPR}, 
  title={Squeeze-and-Excitation Networks}, 
  year={2018},
  volume={},
  number={},
  pages={7132-7141}}

@INPROCEEDINGS{Fan2021-bm,
  author={Fan, Yucai and Yao, Yuhang and Joe-Wong, Carlee},
  booktitle={ICDM}, 
  title={GCN-SE: Attention as Explainability for Node Classification in Dynamic Graphs}, 
  year={2021},
  volume={},
  number={},
  pages={1060-1065}}

@inproceedings{Luo2020-tq,
author = {Luo, Dongsheng and Cheng, Wei and Xu, Dongkuan and Yu, Wenchao and Zong, Bo and Chen, Haifeng and Zhang, Xiang},
title = {Parameterized Explainer for Graph Neural Network},
year = {2020},
booktitle = {NeurIPS},
volume={33},
pages={19620--19631}
}

@inproceedings{Ying2019-zg,
  title={GNNExplainer: generating explanations for graph neural networks},
  author={Ying, Rex and Bourgeois, Dylan and You, Jiaxuan and Zitnik, Marinka and Leskovec, Jure},
  booktitle={NeurIPS},
  pages={9244--9255},
  year={2019}
}

@ARTICLE {Yuan2022-lk,
author = {H. Yuan and H. Yu and S. Gui and S. Ji},
journal = {IEEE T. PAMI},
title = {Explainability in Graph Neural Networks: A Taxonomic Survey},
year = {2023},
volume = {45},
number = {05},
pages = {5782-5799},
}

@article{Goldstein2015-fw,
author = {Alex Goldstein, Adam Kapelner, Justin Bleich and Emil Pitkin},
title = {Peeking Inside the Black Box: Visualizing Statistical Learning With Plots of Individual Conditional Expectation},
journal = {J. of Comput. and Graph. Stat.},
volume = {24},
number = {1},
pages = {44-65},
year = {2015}}

@article{Friedman2001-yz,
author = {Jerome H. Friedman},
title = {{Greedy function approximation: A gradient boosting machine.}},
volume = {29},
journal = {The Annals of Statistics},
number = {5},
pages = {1189 -- 1232},
year = {2001},
}

@inproceedings{Yang2016-xr,
    title = "Hierarchical Attention Networks for Document Classification",
    author = "Yang, Zichao  and
      Yang, Diyi  and
      Dyer, Chris  and
      He, Xiaodong  and
      Smola, Alex  and
      Hovy, Eduard",
    booktitle = "NAACL",
    year = "2016",
    pages = "1480--1489",
}

@article{Vig2019-hx,
  title={Visualizing Attention in Transformer-Based Language Representation Models},
  author={Vig, Jesse},
  journal={arXiv e-prints},
  pages={arXiv--1904},
  year={2019}
}

@inproceedings{Chung2014-ss,
  title={Empirical evaluation of gated recurrent neural networks on sequence modeling},
  author={Chung, Junyoung and Gulcehre, Caglar and Cho, Kyunghyun and Bengio, Yoshua},
  booktitle={NIPS 2014 Workshop on Deep Learning},
  year={2014}
}

@inproceedings{uban-etal-2021-understanding,
    title = "Understanding Patterns of Anorexia Manifestations in Social Media Data with Deep Learning",
    author = "Uban, Ana Sabina  and
      Chulvi, Berta  and
      Rosso, Paolo",
    booktitle = "CLPsych",
    year = "2021",
    pages = "224--236"
}

@article{Uban2021-zw,
title = {An emotion and cognitive based Analysis  of mental health disorders from Social media data},
journal = {Future Generation Computer Syst.},
volume = {124},
pages = {480-494},
year = {2021},
author = {Ana-Sabina Uban and Berta Chulvi and Paolo Rosso},
}

@article{Bahdanau2014-ie,
  title={Neural machine translation by jointly learning to align and translate},
  author={Bahdanau, Dzmitry and Cho, Kyunghyun and Bengio, Yoshua},
  journal={arXiv:1409.0473},
  year={2014}
}

@inproceedings{Vaswani2017-mh,
  author       = {Ashish Vaswani and
                  Noam Shazeer and
                  Niki Parmar and
                  Jakob Uszkoreit and
                  Llion Jones and
                  Aidan N. Gomez and
                  Lukasz Kaiser and
                  Illia Polosukhin},
  title        = {Attention is All you Need},
  booktitle    = {NeurIPS},
  pages        = {5998--6008},
  year         = {2017}
}

@article{Sanh2019-bi,
  title={DistilBERT, a distilled version of BERT: smaller, faster, cheaper and lighter},
  author={Sanh, Victor and Debut, Lysandre and Chaumond, Julien and Wolf, Thomas},
  journal={arXiv:1910.01108},
  pages={arXiv--1910},
  year={2019}
}

@article{Liu2019-vw,
  title={Roberta: A robustly optimized bert pretraining approach},
  author={Liu, Yinhan and Ott, Myle and Goyal, Naman and Du, Jingfei and Joshi, Mandar and Chen, Danqi and Levy, Omer and Lewis, Mike and Zettlemoyer, Luke and Stoyanov, Veselin},
  journal={arXiv:1907.11692},
  year={2019}
}

@ARTICLE{Ragheb2023-nb,
  author={Ragheb, Waleed and Azé, Jérôme and Bringay, Sandra and Servajean, Maximilien},
  journal={IEEE TKDE}, 
  title={Negatively Correlated Noisy Learners for At-Risk User Detection on Social Networks: A Study on Depression, Anorexia, Self-Harm, and Suicide}, 
  year={2023},
  volume={35},
  number={1},
  pages={770-783}}

@inproceedings{Cao2019-fp,
    title = "Latent Suicide Risk Detection on Microblog via Suicide-Oriented Word Embeddings and Layered Attention",
    author = "Cao, Lei  and
      Zhang, Huijun  and
      Feng, Ling  and
      Wei, Zihan  and
      Wang, Xin  and
      Li, Ningyun  and
      He, Xiaohao",
    booktitle = "EMNLP-IJCNLP",
    year = "2019",
    pages = "1718--1728"}

@inproceedings{Matero2019-st,
    title = "Suicide Risk Assessment with Multi-level Dual-Context Language and {BERT}",
    author = "Matero, Matthew  and
      Idnani, Akash  and
      Son, Youngseo  and
      Giorgi, Salvatore  and
      Vu, Huy  and
      Zamani, Mohammad  and
      Limbachiya, Parth  and
      Guntuku, Sharath Chandra  and
      Schwartz, H. Andrew",
    booktitle = "CLPsych",
    year = "2019",
    pages = "39--44"
}

@inproceedings{Van_Aken2020-mc,
  title={Visbert: Hidden-state visualizations for transformers},
  author={Aken, Betty van and Winter, Benjamin and L{\"o}ser, Alexander and Gers, Felix A},
  booktitle={ACM Web Conf.},
  pages={207--211},
  year={2020}
}

@ARTICLE{Naseem2022-vt,
  author={Naseem, Usman and Khushi, Matloob and Kim, Jinman and Dunn, Adam G.},
  journal={IEEE Trans. on Comput. Soc. Syst.}, 
  title={Hybrid Text Representation for Explainable Suicide Risk Identification on Social Media}, 
  year={2022},
  volume={},
  number={},
  pages={1-10}}

@article{Bach2015-vo,
    author = {Bach, Sebastian AND Binder, Alexander AND Montavon, Grégoire AND Klauschen, Frederick AND Müller, Klaus-Robert AND Samek, Wojciech},
    journal = {PLOS ONE},
    title = {On Pixel-Wise Explanations for Non-Linear Classifier Decisions by Layer-Wise Relevance Propagation},
    year = {2015},
    volume = {10},
    pages = {1-46},
    number = {7},

}

@inproceedings{Lundberg2017-xf,
author = {Lundberg, Scott M. and Lee, Su-In},
title = {A Unified Approach to Interpreting Model Predictions},
year = {2017},
booktitle = {NeurIPS},
pages = {4768–4777}
}

@inproceedings{Saxena2022-xp,
author = {Saxena, Chandni and Garg, Muskan and Ansari, Gunjan},
title = {Explainable Causal Analysis of Mental Health on Social Media Data},
year = {2023},
booktitle = {ICONIP},
pages = {172–183}
}

@inproceedings{Ribeiro2016-cn,
author = {Ribeiro, Marco Tulio and Singh, Sameer and Guestrin, Carlos},
title = {"Why Should I Trust You?": Explaining the Predictions of Any Classifier},
year = {2016},
booktitle = {ACM SIGKDD },
pages = {1135–1144},
}

@inproceedings{Amini2020-zk,
  author = {Amini, Hessam and Kosseim, Leila},
  title = {Towards Explainability in Using Deep Learning for the Detection of Anorexia in Social Media},
  booktitle = {NLDB},
  volume = {12089},
  pages = {225--235},
  year = {2020}
}

@inproceedings{Danilevsky2020-cv,
    title = "A Survey of the State of Explainable {AI} for Natural Language Processing",
    author = "Danilevsky, Marina  and
      Qian, Kun  and
      Aharonov, Ranit  and
      Katsis, Yannis  and
      Kawas, Ban  and
      Sen, Prithviraj",
    booktitle = "IJCNLP ",
    year = "2020",
    pages = "447--459",
}

@inproceedings{Ji2021-hu,
    title = "{M}ental{BERT}: Publicly Available Pretrained Language Models for Mental Healthcare",
    author = "Ji, Shaoxiong  and
      Zhang, Tianlin  and
      Ansari, Luna  and
      Fu, Jie  and
      Tiwari, Prayag  and
      Cambria, Erik",
    booktitle = "LREC",
    year = "2022",
    pages = "7184--7190"
}

@inproceedings{Bradley1999AffectiveNF,
  title={Affective Norms for English Words (ANEW): Instruction Manual and Affective Ratings},
  author={Margaret M. Bradley and Peter J. Lang},
  year={1999},
}

@ARTICLE{Nguyen2014-pq,
  author={Nguyen, Thin and Phung, Dinh and Dao, Bo and Venkatesh, Svetha and Berk, Michael},
  journal={IEEE Trans. on Affec. Comp.}, 
  title={Affective and Content Analysis  of Online Depression Communities}, 
  year={2014},
  volume={5},
  number={3},
  pages={217-226}}

@inproceedings{wang-etal-2021-learning,
    title = "Learning Models for Suicide Prediction from Social Media Posts",
    author = "Wang, Ning  and
      Fan, Luo  and
      Shivtare, Yuvraj  and
      Badal, Varsha  and
      Subbalakshmi, Koduvayur  and
      Chandramouli, Rajarathnam  and
      Lee, Ellen",
    booktitle = "CLPsych",
    year = "2021",
    pages = "87--92",
}

@article{Velickovic2017-jd,
  title={Graph attention networks},
  author={Veli{\v{c}}kovi{\'c}, Petar and Cucurull, Guillem and Casanova, Arantxa and Romero, Adriana and Lio, Pietro and Bengio, Yoshua},
  journal={arXiv:1710.10903},
  year={2017}
}

@inproceedings{Naseem2023-xh,
author = {Naseem, Usman and Kim, Jinman and Khushi, Matloob and Dunn, Adam},
title = {Graph-Based Hierarchical Attention Network for Suicide Risk Detection on Social Media},
year = {2023},
booktitle = {ACM Web Conf. },
pages = {995–1003},
}

@inproceedings{Sawhney2021-wd,
    title = "Suicide Ideation Detection via Social and Temporal User Representations using Hyperbolic Learning",
    author = "Sawhney, Ramit  and
      Joshi, Harshit  and
      Shah, Rajiv Ratn  and
      Flek, Lucie",
    booktitle = "NAACL",
    year = "2021",
    pages = "2176--2190",
}

@inproceedings{sawhney-etal-2022-risk,
    title = "A Risk-Averse Mechanism for Suicidality Assessment on Social Media",
    author = "Sawhney, Ramit  and
      Neerkaje, Atula  and
      Gaur, Manas",
    booktitle = "ACL",
    year = "2022",
    pages = "628--635"
}

@inproceedings{Sawhney2020-en,
    title = "A Time-Aware Transformer Based Model for Suicide Ideation Detection on Social Media",
    author = "Sawhney, Ramit  and
      Joshi, Harshit  and
      Gandhi, Saumya  and
      Shah, Rajiv Ratn",
    booktitle = "EMNLP",
    year = "2020",
    pages = "7685--7697",
}

@inproceedings{10.1145/3543507.3583867,
author = {Wu, Jiageng and Wu, Xian and Hua, Yining and Lin, Shixu and Zheng, Yefeng and Yang, Jie},
title = {Exploring Social Media for Early Detection of Depression in COVID-19 Patients},
year = {2023},
booktitle = {ACM Web Conf.},
pages = {3968–3977}
}

@ARTICLE{Sanchez2020-wf,
  title    = "Social media recruitment for mental health research: A systematic
              review",
  author   = "Sanchez, Catherine and Grzenda, Adrienne and Varias, Andrea and
              Widge, Alik S and Carpenter, Linda L and McDonald, William M and
              Nemeroff, Charles B and Kalin, Ned H and Martin, Glenn and Tohen,
              Mauricio and Filippou-Frye, Maria and Ramsey, Drew and Linos,
              Eleni and Mangurian, Christina and Rodriguez, Carolyn I",
  journal  = "Compr. Psychiatry",
  volume   =  103,
  pages    = "152197",
  year     =  2020
}

@ARTICLE{Jia2022-hp,
  title     = "The Role of Explainability in Assuring Safety of Machine
               Learning in Healthcare",
  author    = "Jia, Yan and McDermid, John and Lawton, Tom and Habli, Ibrahim",
  journal   = "IEEE T. on Emerging Topics in Computing",
  volume    =  10,
  number    =  4,
  pages     = "1746--1760",
  year      =  2022
}

@ARTICLE{Tlachac2020-qw,
  author={Tlachac, ML and Rundensteiner, Elke},
  journal={IEEE J. of Biomed. and Health Inform.}, 
  title={Screening For Depression With Retrospectively Harvested Private Versus Public Text}, 
  year={2020},
  volume={24},
  number={11},
  pages={3326-3332}}

@article{Greenwell2018-yl,
  title={A simple and effective model-based variable importance measure},
  author={Greenwell, Brandon M and Boehmke, Bradley C and McCarthy, Andrew J},
  journal={arXiv:1805.04755},
  year={2018}
}

@ARTICLE{Lecun1998-pu,
  author={Lecun, Y. and Bottou, L. and Bengio, Y. and Haffner, P.},
  journal={Proc. of the IEEE}, 
  title={Gradient-based learning applied to document recognition}, 
  year={1998},
  volume={86},
  number={11},
  pages={2278-2324}}

@inproceedings{Chami2019-sc,
 author = {Chami, Ines and Ying, Zhitao and R\'{e}, Christopher and Leskovec, Jure},
 booktitle = {NeurIPS},
 editor = {H. Wallach and H. Larochelle and A. Beygelzimer and F. d\textquotesingle Alch\'{e}-Buc and E. Fox and R. Garnett},
 pages = {},
 title = {Hyperbolic Graph Convolutional Neural Networks},
 volume = {32},
 year = {2019}
}

@inproceedings{Losada2019-cm,
author = {Losada, David E. and Crestani, Fabio and Parapar, Javier},
title = {Overview of ERisk 2019 Early Risk Prediction on the Internet},
year = {2019},
booktitle = {CLEF},
pages = {340–357},
}

@article{10.1145/3359169,
author = {Pendse, Sachin R. and Niederhoffer, Kate and Sharma, Amit},
title = {Cross-Cultural Differences in the Use of Online Mental Health Support Forums},
year = {2019},
volume = {3},
number = {CSCW},
journal = {Proc. ACM Hum.-Comput. Interact.},
articleno = {67},
}

@article{gopalkrishnan2018cultural,
  title={Cultural diversity and mental health: Considerations for policy and practice},
  author={Gopalkrishnan, Narayan},
  journal={Frontiers in public health},
  volume={6},
  pages={179},
  year={2018},
}

@article{10.1145/3359249,
author = {Chancellor, Stevie and Baumer, Eric P. S. and De Choudhury, Munmun},
title = {Who is the "Human" in Human-Centered Machine Learning: The Case of Predicting Mental Health from Social Media},
year = {2019},
volume = {3},
number = {CSCW},
journal = {Proc. ACM Hum.-Comput. Interact.},
articleno = {147},
}

@article{MORLEY2020113172,
title = {The ethics of AI in health care: A mapping review},
journal = {Social Science and Medicine},
volume = {260},
pages = {113172},
year = {2020},
author = {Jessica Morley and Caio C.V. Machado and Christopher Burr and Josh Cowls and Indra Joshi and Mariarosaria Taddeo and Luciano Floridi},

}

@article{chancellor2020methods,
  title={Methods in predictive techniques for mental health status on social media: a critical review},
  author={Chancellor, Stevie and De Choudhury, Munmun},
  journal={NPJ digital medicine},
  volume={3},
  number={1},
  pages={43},
  year={2020}
}

@article{schonning2020social,
  title={Social media use and mental health and well-being among adolescents--a scoping review},
  author={Sch{\o}nning, Viktor and Hjetland, Gunnhild Johnsen and Aar{\o}, Leif Edvard and Skogen, Jens Christoffer},
  journal={Frontiers in psychology},
  volume={11},
  pages={542107},
  year={2020}
}

@ARTICLE{10108975,
  author={Hasib, Khan Md and Islam, Md Rafiqul and Sakib, Shadman and Akbar, Md. Ali and Razzak, Imran and Alam, Mohammad Shafiul},
  journal={IEEE Transactions on Computational Social Systems}, 
  title={Depression Detection From Social Networks Data Based on Machine Learning and Deep Learning Techniques: An Interrogative Survey}, 
  year={2023},
  volume={10},
  number={4},
  pages={1568-1586}}

@ARTICLE{9568643,
  author={Xu, Xinyuan},
  journal={IEEE Transactions on Computational Social Systems}, 
  title={Detecting Suicide Ideation in the Online Environment: A Survey of Methods and Challenges}, 
  year={2022},
  volume={9},
  number={3},
  pages={679-687}}

@ARTICLE{9733425,
  author={Ansari, Luna and Ji, Shaoxiong and Chen, Qian and Cambria, Erik},
  journal={IEEE Transactions on Computational Social Systems}, 
  title={Ensemble Hybrid Learning Methods for Automated Depression Detection}, 
  year={2023},
  volume={10},
  number={1},
  pages={211-219}}

@ARTICLE{10241281,
  author={Anshul, Ashutosh and Pranav, Gumpili Sai and Rehman, Mohammad Zia Ur and Kumar, Nagendra},
  journal={IEEE Transactions on Computational Social Systems}, 
  title={A Multimodal Framework for Depression Detection During COVID-19 via Harvesting Social Media}, 
  year={2024},
  volume={11},
  number={2},
  pages={2872-2888}}

@article{ZHANG2024111650,
title = {Enhancing user sequence representation with cross-view collaborative learning for depression detection on Sina Weibo},
journal = {Knowledge-Based Systems},
volume = {293},
pages = {111650},
year = {2024},
doi = {https://doi.org/10.1016/j.knosys.2024.111650},
author = {Zhenwen Zhang and Zepeng Li and Jianghong Zhu and Zhihua Guo and Bin Shi and Bin Hu},
}

@article{rehm2019global,
  title={Global burden of disease and the impact of mental and addictive disorders},
  author={Rehm, J{\"u}rgen and Shield, Kevin D},
  journal={Current psychiatry reports},
  volume={21},
  pages={1--7},
  year={2019},
  publisher={Springer}
}

@article{thornicroft2017undertreatment,
  title={Undertreatment of people with major depressive disorder in 21 countries},
  author={Thornicroft, Graham and Chatterji, Somnath and Evans-Lacko, Sara and Gruber, Michael and Sampson, Nancy and Aguilar-Gaxiola, Sergio and Al-Hamzawi, Ali and Alonso, Jordi and Andrade, Laura and Borges, Guilherme and others},
  journal={The British Journal of Psychiatry},
  volume={210},
  number={2},
  pages={119--124},
  year={2017},
  publisher={Cambridge University Press}
}

@article{trautmann2016economic,
  title={The economic costs of mental disorders: Do our societies react appropriately to the burden of mental disorders?},
  author={Trautmann, Sebastian and Rehm, J{\"u}rgen and Wittchen, Hans-Ulrich},
  journal={EMBO reports},
  volume={17},
  number={9},
  pages={1245--1249},
  year={2016}
}

@article{henderson2020mental,
  title={Mental illness stigma after a decade of Time to Change England: inequalities as targets for further improvement},
  author={Henderson, Claire and Potts, Laura and Robinson, Emily J},
  journal={European journal of public health},
  volume={30},
  number={3},
  pages={497--503},
  year={2020},
  publisher={Oxford University Press}
}

@article{alho2024transmission,
  title={Transmission of mental disorders in adolescent peer networks},
  author={Alho, Jussi and Gutvilig, Mai and Niemi, Ripsa and Komulainen, Kaisla and B{\"o}ckerman, Petri and Webb, Roger T and Elovainio, Marko and Hakulinen, Christian},
  journal={JAMA psychiatry},
  volume={81},
  number={9},
  pages={882--888},
  year={2024},
  publisher={American Medical Association}
}

@article{kiuru2012depression,
  title={Is depression contagious? A test of alternative peer socialization mechanisms of depressive symptoms in adolescent peer networks},
  author={Kiuru, Noona and Burk, William J and Laursen, Brett and Nurmi, Jari-Erik and Salmela-Aro, Katariina},
  journal={Journal of Adolescent Health},
  volume={50},
  number={3},
  pages={250--255},
  year={2012},
  publisher={Elsevier}
}

@article{kelly2025interpretable,
  title={An Interpretable Model With Probabilistic Integrated Scoring for Mental Health Treatment Prediction: Design Study},
  author={Kelly, Anthony and Jensen, Esben Kjems and Grua, Eoin Martino and Mathiasen, Kim and Van de Ven, Pepijn},
  journal={JMIR Medical Informatics},
  volume={13},
  pages={e64617},
  year={2025},
  publisher={JMIR Publications Toronto, Canada}
}

@article{LIU2025529,
title = {Developing an interpretable machine learning model for screening depression in older adults with functional disability},
journal = {Journal of Affective Disorders},
volume = {379},
pages = {529-539},
year = {2025},
issn = {0165-0327},
author = {Deyan Liu and Yuge Tian and Min Liu and Shangjian Yang},
}

@article{kerz2023toward,
  title={Toward explainable AI (XAI) for mental health detection based on language behavior},
  author={Kerz, Elma and Zanwar, Sourabh and Qiao, Yu and Wiechmann, Daniel},
  journal={Frontiers in psychiatry},
  volume={14},
  pages={1219479},
  year={2023},
  publisher={Frontiers Media SA}
}

@article{alghazzawi2025explainable,
  title={Explainable AI-based suicidal and non-suicidal ideations detection from social media text with enhanced ensemble technique},
  author={Alghazzawi, Daniyal and Ullah, Hayat and Tabassum, Naila and Badri, Sahar K and Asghar, Muhammad Zubair},
  journal={Scientific Reports},
  volume={15},
  number={1},
  pages={1111},
  year={2025},
  publisher={Nature Publishing Group UK London}
}

@inproceedings{deyoung-etal-2020-eraser,
    title = "{ERASER}: {A} Benchmark to Evaluate Rationalized {NLP} Models",
    author = "DeYoung, Jay  and
      Jain, Sarthak  and
      Rajani, Nazneen Fatema  and
      Lehman, Eric  and
      Xiong, Caiming  and
      Socher, Richard  and
      Wallace, Byron C.",
    editor = "Jurafsky, Dan  and
      Chai, Joyce  and
      Schluter, Natalie  and
      Tetreault, Joel",
    booktitle = "Proceedings of the 58th ACL",
    month = jul,
    year = "2020",
    address = "Online",
    publisher = "Association for Computational Linguistics",
    pages = "4443--4458",
}

@inproceedings{holland2021robustness,
  title={Robustness and scalability under heavy tails, without strong convexity},
  author={Holland, Matthew},
  booktitle={Int. Conf. on Artificial Intelligence and Statistics},
  pages={865--873},
  year={2021},
  organization={PMLR}
}

@inproceedings{wiegreffe2019attention,
  title={Attention is not not Explanation},
  author={Wiegreffe, Sarah and Pinter, Yuval},
  booktitle={Proc. Conf. on EMNLP-IJCNLP},
  pages={11--20},
  year={2019}
}

@INPROCEEDINGS{9671639,
  author={Liu, Shengzhong and Le, Franck and Chakraborty, Supriyo and Abdelzaher, Tarek},
  booktitle={2021 IEEE International Conference on Big Data (Big Data)}, 
  title={On Exploring Attention-based Explanation for Transformer Models in Text Classification}, 
  year={2021},
  volume={},
  number={},
  pages={1193-1203}}

@article{hoover2019exbert,
  title={exbert: A visual analysis tool to explore learned representations in transformers models},
  author={Hoover, Benjamin and Strobelt, Hendrik and Gehrmann, Sebastian},
  journal={arXiv preprint arXiv:1910.05276},
  year={2019}
}

@inproceedings{hu2023seat,
  title={Seat: stable and explainable attention},
  author={Hu, Lijie and Liu, Yixin and Liu, Ninghao and Huai, Mengdi and Sun, Lichao and Wang, Di},
  booktitle={AAAI},
  volume={37},
  number={11},
  pages={12907--12915},
  year={2023}
}

@inproceedings{10.1145/3589334.3648137,
author = {Yang, Kailai and Zhang, Tianlin and Kuang, Ziyan and Xie, Qianqian and Huang, Jimin and Ananiadou, Sophia},
title = {MentaLLaMA: Interpretable Mental Health Analysis on Social Media with Large Language Models},
year = {2024},
isbn = {9798400701719},
publisher = {Association for Computing Machinery},
address = {New York, NY, USA},
booktitle = {Proceedings of the ACM Web Conference 2024},
pages = {4489–4500},
numpages = {12},
location = {Singapore, Singapore},
series = {WWW '24}
}

@article{bang2023multitask,
  title={A multitask, multilingual, multimodal evaluation of chatgpt on reasoning, hallucination, and interactivity},
  author={Bang, Yejin and Cahyawijaya, Samuel and Lee, Nayeon and Dai, Wenliang and Su, Dan and Wilie, Bryan and Lovenia, Holy and Ji, Ziwei and Yu, Tiezheng and Chung, Willy and others},
  journal={arXiv preprint arXiv:2302.04023},
  year={2023}
}

@article{deshpande2023toxicity,
  title={Toxicity in chatgpt: Analyzing persona-assigned language models},
  author={Deshpande, Ameet and Murahari, Vishvak and Rajpurohit, Tanmay and Kalyan, Ashwin and Narasimhan, Karthik},
  journal={arXiv preprint arXiv:2304.05335},
  year={2023}
}

@article{yin2023large,
  title={Do large language models know what they don't know?},
  author={Yin, Zhangyue and Sun, Qiushi and Guo, Qipeng and Wu, Jiawen and Qiu, Xipeng and Huang, Xuanjing},
  journal={arXiv preprint arXiv:2305.18153},
  year={2023}
}

@inproceedings{xie2023adaptive,
  title={Adaptive chameleon or stubborn sloth: Revealing the behavior of large language models in knowledge conflicts},
  author={Xie, Jian and Zhang, Kai and Chen, Jiangjie and Lou, Renze and Su, Yu},
  booktitle={The Twelfth International Conference on Learning Representations},
  year={2023}
}

@article{wang2023resolving,
  title={Resolving knowledge conflicts in large language models},
  author={Wang, Yike and Feng, Shangbin and Wang, Heng and Shi, Weijia and Balachandran, Vidhisha and He, Tianxing and Tsvetkov, Yulia},
  journal={arXiv preprint arXiv:2310.00935},
  year={2023}
}

@article{ju2024large,
  title={How large language models encode context knowledge? a layer-wise probing study},
  author={Ju, Tianjie and Sun, Weiwei and Du, Wei and Yuan, Xinwei and Ren, Zhaochun and Liu, Gongshen},
  journal={arXiv preprint arXiv:2402.16061},
  year={2024}
}

@article{thieme_rev,
author = {Thieme, Anja and Belgrave, Danielle and Doherty, Gavin},
title = {Machine Learning in Mental Health: A Systematic Review of the HCI Literature to Support the Development of Effective and Implementable ML Systems},
year = {2020},
issue_date = {October 2020},
publisher = {Association for Computing Machinery},
address = {New York, NY, USA},
volume = {27},
number = {5},
issn = {1073-0516},
journal = {ACM Trans. Comput.-Hum. Interact.},
month = aug,
articleno = {34},
numpages = {53}
}

@article{belcastro2025detecting,
  title={Detecting mental disorder on social media: a ChatGPT-augmented explainable approach},
  author={Belcastro, Loris and Cantini, Riccardo and Marozzo, Fabrizio and Talia, Domenico and Trunfio, Paolo},
  journal={Online Social Networks and Media},
  volume={48},
  pages={100321},
  year={2025},
  publisher={Elsevier}
}

@article{al2025effective,
  title={Effective depression detection and interpretation: Integrating machine learning, deep learning, language models, and explainable AI},
  author={Al Masud, Gazi Hasan and Shanto, Rejaul Islam and Sakin, Ishmam and Kabir, Muhammad Rafsan},
  journal={Array},
  volume={25},
  pages={100375},
  year={2025},
  publisher={Elsevier}
}

@article{Bao2024,
  author    = {Bao, Eliseo and Pérez, Anxo and Parapar, Javier},
  title     = {Explainable depression symptom detection in social media},
  journal   = {Health Information Science and Systems},
  year      = {2024},
  volume    = {12},
  number    = {1},
  pages     = {47}
}

@InProceedings{pmlr-v284-mahdavinejad25a,
  title = 	 {Towards Explainable Depression Detection: A Neurosymbolic Approach to Uncover Social Media Signals with Generative AI},
  author =       {Mahdavinejad, Mohammad Saeid and Adibi, Peyman and Monajemi, Amirhassan and Hitzler, Pascal},
  booktitle = 	 {Proceedings of The 19th International Conference on Neurosymbolic Learning and Reasoning},
  pages = 	 {830--853},
  year = 	 {2025},
  editor = 	 {H. Gilpin, Leilani and Giunchiglia, Eleonora and Hitzler, Pascal and van Krieken, Emile},
  volume = 	 {284},
  series = 	 {Proceedings of Machine Learning Research},
  month = 	 {08--10 Sep},
  publisher =    {PMLR}
}

@article{hameed2025explainable,
  title={Explainable AI-driven depression detection from social media using natural language processing and black box machine learning models},
  author={Hameed, Sidra and Nauman, Muhammad and Akhtar, Nadeem and Fayyaz, Muhammad AB and Nawaz, Raheel},
  journal={Frontiers in Artificial Intelligence},
  volume={8},
  pages={1627078},
  year={2025},
  publisher={Frontiers Media SA}
}

@article{Lamba2026,
  author    = {Lamba, Kamini and Rani, Shalli and Shabaz, Mohammad},
  title     = {Explainable machine learning for mental health prediction from social media behavior: a nested cross-validation study with SHAP and LIME interpretability},
  journal   = {Discover Mental Health},
  year      = {2026},
  volume    = {6},
  number    = {1},
  pages     = {25}
}

@inproceedings{poswiata-perelkiewicz-2022-opi,
    title = "{OPI}@{LT}-{EDI}-{ACL}2022: Detecting Signs of Depression from Social Media Text using {R}o{BERT}a Pre-trained Language Models",
    author = "Po{\'s}wiata, Rafa{\l}  and
      Pere{\l}kiewicz, Micha{\l}",
    editor = "Chakravarthi, Bharathi Raja  and
      Bharathi, B  and
      McCrae, John P  and
      Zarrouk, Manel  and
      Bali, Kalika  and
      Buitelaar, Paul",
    booktitle = "Proceedings of the Second Workshop on Language Technology for Equality, Diversity and Inclusion",
    month = may,
    year = "2022",
    address = "Dublin, Ireland",
    publisher = "Association for Computational Linguistics"
}

@inproceedings{giorgi2022correcting,
  title={Correcting sociodemographic selection biases for population prediction from social media},
  author={Giorgi, Salvatore and Lynn, Veronica E and Gupta, Keshav and Ahmed, Farhan and Matz, Sandra and Ungar, Lyle H and Schwartz, H Andrew},
  booktitle={ICWSM},
  volume={16},
  pages={228--240},
  year={2022}
}

@article{olteanu2019social,
  title={Social data: Biases, methodological pitfalls, and ethical boundaries},
  author={Olteanu, Alexandra and Castillo, Carlos and Diaz, Fernando and K{\i}c{\i}man, Emre},
  journal={Frontiers in big data},
  volume={2},
  pages={13},
  year={2019},
  publisher={Frontiers Media SA}
}

@online{AIActExplorer2024,
  author       = {{Future of Life Institute}},
  title        = {{AI Act Explorer}},
  year         = {2024},
  url          = {https://artificialintelligenceact.eu/ai-act-explorer/},
  note         = {Accessed: 2026-03-04}
}

@misc{WHO_LMM_2025,
  author       = {{World Health Organization}},
  title        = {{Ethics and governance of artificial intelligence for health: Guidance on large multi-modal models}},
  year         = {2025},
  month        = mar,
  howpublished = {\url{https://www.who.int/publications/i/item/9789240084759}},
  note         = {Published 25 March 2025, 98 pages, ISBN: 978-92-4-008475-9}
}

@article{neiders2025ethical,
  title={Ethical and social issues in prediction of risk of severe mental illness: a scoping review and thematic analysis},
  author={Neiders, Ivars and Me{\v{z}}inska, Signe and van Haren, Neeltje EM},
  journal={BMC psychiatry},
  volume={25},
  number={1},
  pages={501},
  year={2025},
  publisher={Springer}
}

@article{lane2022towards,
  title={Towards personalised predictive psychiatry in clinical practice: an ethical perspective},
  author={Lane, Natalie and Broome, Matthew},
  journal={The British Journal of Psychiatry},
  volume={220},
  number={4},
  pages={172--174},
  year={2022},
  publisher={Cambridge University Press}
}

@inproceedings{yoon2022d,
  title={D-vlog: Multimodal vlog dataset for depression detection},
  author={Yoon, Jeewoo and Kang, Chaewon and Kim, Seungbae and Han, Jinyoung},
  booktitle={AAAI},
  volume={36},
  number={11},
  pages={12226--12234},
  year={2022}
}

@article{min2023detecting,
  title={Detecting depression on video logs using audiovisual features},
  author={Min, Kyungeun and Yoon, Jeewoo and Kang, Migyeong and Lee, Daeun and Park, Eunil and Han, Jinyoung},
  journal={Humanities and Social Sciences Communications},
  volume={10},
  number={1},
  pages={1--8},
  year={2023},
  publisher={Palgrave}
}

@article{bereska2024mechanistic,
  title={Mechanistic interpretability for AI safety--a review},
  author={Bereska, Leonard and Gavves, Efstratios},
  journal={arXiv preprint arXiv:2404.14082},
  year={2024}
}

@article{wang2022interpretability,
  title={Interpretability in the wild: a circuit for indirect object identification in gpt-2 small},
  author={Wang, Kevin and Variengien, Alexandre and Conmy, Arthur and Shlegeris, Buck and Steinhardt, Jacob},
  journal={arXiv preprint arXiv:2211.00593},
  year={2022}
}

@article{yang2025retrieval,
  title={Retrieval-augmented generation for generative artificial intelligence in health care},
  author={Yang, Rui and Ning, Yilin and Keppo, Emilia and Liu, Mingxuan and Hong, Chuan and Bitterman, Danielle S and Ong, Jasmine Chiat Ling and Ting, Daniel Shu Wei and Liu, Nan},
  journal={Npj health systems},
  volume={2},
  number={1},
  pages={2},
  year={2025},
  publisher={Nature Publishing Group UK London}
}

@article{klesel2025retrieval,
  title={Retrieval-augmented generation (rag) m. klesel, hf wittmann},
  author={Klesel, Michael and Wittmann, H Felix},
  journal={Business \& Information Systems Engineering},
  volume={67},
  number={4},
  pages={551--561},
  year={2025},
  publisher={Springer}
}

@article{kharitonova2025incorporating,
  title={Incorporating evidence into mental health Q\&A: a novel method to use generative language models for validated clinical content extraction},
  author={Kharitonova, Ksenia and P{\'e}rez-Fern{\'a}ndez, David and Guti{\'e}rrez-Hernando, Javier and Guti{\'e}rrez-Fandi{\~n}o, Asier and Callejas, Zoraida and Griol, David},
  journal={Behaviour \& Information Technology},
  volume={44},
  number={10},
  pages={2333--2350},
  year={2025},
  publisher={Taylor \& Francis}
}

@article{ge2025survey,
  title={Survey and Experiments on Mental Disorder Detection via Social Media: From Large Language Models and RAG to Agents},
  author={Ge, Zhuohan and Li, Darian and Wang, Yubo and Hu, Nicole and Zhu, Xinyi and Li, Haoyang and Zhang, Xin and Zhang, Mingtao and Qi, Shihao and Xu, Yuming and others},
  journal={arXiv preprint arXiv:2504.02800},
  year={2025}
}

@inproceedings{mothilal2020explaining,
  title={Explaining machine learning classifiers through diverse counterfactual explanations},
  author={Mothilal, Ramaravind K and Sharma, Amit and Tan, Chenhao},
  booktitle={Proceedings of the 2020 conference on fairness, accountability, and transparency},
  pages={607--617},
  year={2020}
}

@article{ali2025artificial,
  title={Artificial intelligence for mental health: A narrative review of applications, challenges, and future directions in digital health},
  author={Ali, Maisam and Ali, Shahid and Abbas, Qaiser and Abbas, Zeeshan and Lee, Seung Won},
  journal={Digital Health},
  volume={11},
  pages={20552076251395548},
  year={2025},
  publisher={SAGE Publications Sage UK: London, England}
}

@article{mertes2022ganterfactual,
  title={Ganterfactual—counterfactual explanations for medical non-experts using generative adversarial learning},
  author={Mertes, Silvan and Huber, Tobias and Weitz, Katharina and Heimerl, Alexander and Andr{\'e}, Elisabeth},
  journal={Frontiers in artificial intelligence},
  volume={5},
  pages={825565},
  year={2022},
  publisher={Frontiers Media SA}
}

@article{qin2025explainable,
  title={Explainable Counterfactual Reasoning in Depression Medication Selection at Multi-Levels (Personalized and Population)},
  author={Qin, Xinyu and Chignell, Mark H and Greifenberger, Alexandria and Lokuge, Sachinthya and Toumeh, Elssa and Sternat, Tia and Katzman, Martin and Wang, Lu},
  journal={arXiv preprint arXiv:2508.17207},
  year={2025}
}

@article{kandala2025explainability,
  title={From Explainability to Action: A Generative Operational Framework for Integrating XAI in Clinical Mental Health Screening},
  author={Kandala, Ratna and Moharir, Akshata Kishore and Nayak, Divya Arvinda},
  journal={arXiv preprint arXiv:2510.13828},
  year={2025}
}

@article{lee2025prompt,
  title={A prompt framework for enhancing LLM-based explainability of medical machine learning models: an intensive care unit application},
  author={Lee, Sujung and Cho, Won Ik and Lee, Youngrong and Kim, Duck Ju and Nam, Kyeng Hyun and Lee, Sangmin and Suh, Jungyo and Ko, Taehoon},
  journal={BMC Medical Informatics and Decision Making},
  volume={25},
  number={1},
  pages={430},
  year={2025},
  publisher={Springer}
}

@inproceedings{zytek2024explingo,
  title={Explingo: Explaining ai predictions using large language models},
  author={Zytek, Alexandra and Pido, Sara and Alnegheimish, Sarah and Berti-Equille, Laure and Veeramachaneni, Kalyan},
  booktitle={2024 IEEE International Conference on Big Data (BigData)},
  pages={1197--1208},
  year={2024},
  organization={IEEE}
}

@article{kim2503medical,
  title={Medical hallucinations in foundation models and their impact on healthcare. arXiv 2025},
  author={Kim, Y and Jeong, H and Chen, S and Li, SS and Lu, M and Alhamoud, K and Mun, J and Grau, C and Jung, M and Gameiro, R and others},
  journal={arXiv preprint arXiv:2503.05777}
}

@article{rahman2024depressionemo,
  title={DepressionEmo: A novel dataset for multilabel classification of depression emotions},
  author={Rahman, Abu Bakar Siddiqur and Ta, Hoang-Thang and Najjar, Lotfollah and Azadmanesh, Azad and G{\"o}nul, Ali Saffet},
  journal={Journal of Affective Disorders},
  volume={366},
  pages={445--458},
  year={2024},
  publisher={Elsevier}
}

@inproceedings{alhamed2024classifying,
  title={Classifying social media users before and after depression diagnosis via their language usage: A dataset and study},
  author={Alhamed, Falwah and Ive, Julia and Specia, Lucia},
  booktitle={Proceedings of the 2024 Joint International Conference on Computational Linguistics, Language Resources and Evaluation (LREC-COLING 2024)},
  pages={3250--3260},
  year={2024}
}

@article{beniwal2024hybrid,
  title={A hybrid BERT-CNN approach for depression detection on social media using multimodal data},
  author={Beniwal, Rohit and Saraswat, Pavi},
  journal={The Computer Journal},
  volume={67},
  number={7},
  pages={2453--2472},
  year={2024},
  publisher={Oxford University Press}
}

@inproceedings{romero2024mentalriskes,
  title={MentalRiskES: A new corpus for early detection of mental disorders in Spanish},
  author={Romero, Alba Mar{\'\i}a M{\'a}rmol and Moreno-Mu{\~n}oz, Adri{\'a}n and Plaza-Del-Arco, Flor Miriam and Molina-Gonz{\'a}lez, M Dolores and Montejo-R{\'a}ez, Arturo},
  booktitle={Proceedings of the 2024 Joint International Conference on Computational Linguistics, Language Resources and Evaluation (LREC-COLING 2024)},
  pages={11204--11214},
  year={2024}
}

@article{ghosh2023attention,
  title={An attention-based hybrid architecture with explainability for depressive social media text detection in Bangla},
  author={Ghosh, Tapotosh and Al Banna, Md Hasan and Al Nahian, Md Jaber and Uddin, Mohammed Nasir and Kaiser, M Shamim and Mahmud, Mufti},
  journal={Expert Systems with Applications},
  volume={213},
  pages={119007},
  year={2023},
  publisher={Elsevier}
}

@article{li2023mha,
  title={MHA: a multimodal hierarchical attention model for depression detection in social media},
  author={Li, Zepeng and An, Zhengyi and Cheng, Wenchuan and Zhou, Jiawei and Zheng, Fang and Hu, Bin},
  journal={Health information science and systems},
  volume={11},
  number={1},
  pages={6},
  year={2023},
  publisher={Springer}
}

@article{guo2023leveraging,
  title={Leveraging domain knowledge to improve depression detection on Chinese social media},
  author={Guo, Zhihua and Ding, Nengneng and Zhai, Minyu and Zhang, Zhenwen and Li, Zepeng},
  journal={IEEE Transactions on Computational Social Systems},
  volume={10},
  number={4},
  pages={1528--1536},
  year={2023},
  publisher={IEEE}
}

@inproceedings{tsakalidis2022identifying,
  title={Identifying moments of change from longitudinal user text},
  author={Tsakalidis, Adam and Nanni, Federico and Hills, Anthony and Chim, Jenny and Song, Jiayu and Liakata, Maria},
  booktitle={Proceedings of the 60th Annual Meeting of the Association for Computational Linguistics (Volume 1: Long Papers)},
  pages={4647--4660},
  year={2022}
}

@inproceedings{agarwal2025redepress,
  title={ReDepress: A Cognitive Framework for Detecting Depression Relapse from Social Media},
  author={Agarwal, Aakash Kumar and Bhattacharjee, Saprativa and Rastogi, Mauli and Jacob, Jemima S and Banerjee, Biplab and Gupta, Rashmi and Bhattacharyya, Pushpak},
  booktitle={Proc. EMNLP},
  pages={34652--34670},
  year={2025}
}

@article{al2018video,
  title={Video-based depression level analysis by encoding deep spatiotemporal features},
  author={Al Jazaery, Mohamad and Guo, Guodong},
  journal={IEEE Transactions on Affective Computing},
  volume={12},
  number={1},
  pages={262--268},
  year={2018},
  publisher={IEEE}
}

@article{li2025depressinstruct,
  title={DepressInstruct: Instruction Tuning of Large Speech-Language Models for Depression Detection},
  author={Li, Dongdong and Ding, Li and Wang, Zhe and Zhao, Ke},
  journal={Information Fusion},
  pages={104077},
  year={2025},
  publisher={Elsevier}
}

@article{zhang2026sounding,
  title={Sounding Depressed? Personalized Deep Learning Model for Depression Detection from Speech and Text},
  author={Zhang, Haibo and Liu, Zhenyu and Yuan, Jiaqian and Li, Gang and Ding, Zhijie and Hu, Bin},
  journal={IEEE Transactions on Multimedia},
  year={2026},
  publisher={IEEE}
}

\end{sloppypar}

\end{document}